\documentclass[10pt,twocolumn,letterpaper]{article}

\usepackage[pagenumbers]{cvpr} 

\usepackage{graphicx}
\usepackage{amsmath}
\usepackage{amssymb}
\usepackage{booktabs,threeparttable}
\usepackage{colortbl}
\usepackage[export]{adjustbox}
\usepackage{caption}
\usepackage{multirow}

\usepackage[pagebackref,breaklinks,colorlinks]{hyperref}

\usepackage[capitalize]{cleveref}
\crefname{section}{Sec.}{Secs.}
\Crefname{section}{Section}{Sections}
\Crefname{table}{Table}{Tables}
\crefname{table}{Tab.}{Tabs.}

\def\cvprPaperID{****} 
\def\confName{CCCC}
\def\confYear{YYYY}

\begin{document}
		
	\title{YOLOv9: Learning What You Want to Learn \\ Using Programmable Gradient Information}

	\author{
		\vspace{-24pt} \\
		Chien-Yao Wang$^{1, 2}$, I-Hau Yeh$^{2}$, and Hong-Yuan Mark Liao$^{1, 2, 3}$ \\
		$^{1}$Institute of Information Science, Academia Sinica, Taiwan \\
		$^{2}$National Taipei University of Technology, Taiwan \\
		$^{3}$Department of Information and Computer Engineering, Chung Yuan Christian University, Taiwan \\
		{\tt\small kinyiu@iis.sinica.edu.tw, ihyeh@emc.com.tw, and liao@iis.sinica.edu.tw}
		\vspace{-20pt}
	}
	
	\maketitle
	
	\begin{abstract}
		
	\vspace{-8pt}
	
	Today's deep learning methods focus on how to design the most appropriate objective functions so that the prediction results of the model can be closest to the ground truth. Meanwhile, an appropriate architecture that can facilitate acquisition of enough information for prediction has to be designed. Existing methods ignore a fact that when input data undergoes layer-by-layer feature extraction and spatial transformation, large amount of information will be lost. This paper will delve into the important issues of data loss when data is transmitted through deep networks, namely information bottleneck and reversible functions. We proposed the concept of programmable gradient information (PGI) to cope with the various changes required by deep networks to achieve multiple objectives. PGI can provide complete input information for the target task to calculate objective function, so that reliable gradient information can be obtained to update network weights. In addition, a new lightweight network architecture -- Generalized Efficient Layer Aggregation Network (GELAN), based on gradient path planning is designed. GELAN's architecture confirms that PGI has gained superior results on lightweight models. We verified the proposed GELAN and PGI on MS COCO dataset based object detection. The results show that GELAN only uses conventional convolution operators to achieve better parameter utilization than the state-of-the-art methods developed based on depth-wise convolution. PGI can be used for variety of models from lightweight to large. It can be used to obtain complete information, so that train-from-scratch models can achieve better results than state-of-the-art models pre-trained using large datasets, the comparison results are shown in Figure~\ref{fig:sota}. The source codes are at: \url{https://github.com/WongKinYiu/yolov9}.
	
	\vspace{-16pt}
		
	\end{abstract}

	\begin{figure}[t]
		\begin{center}
			\includegraphics[width=.91\linewidth]{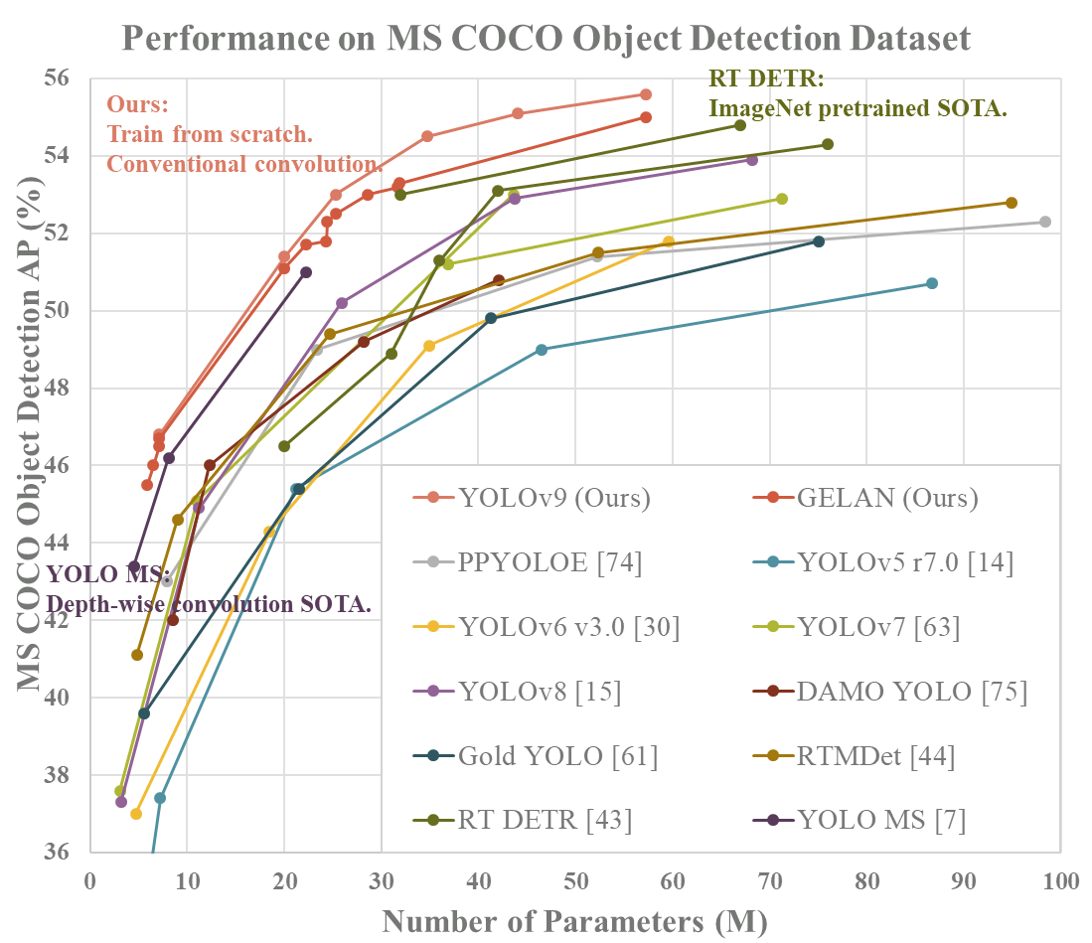}
		\end{center}
		\vspace{-10pt}
		\caption{Comparisons of the real-time object detecors on MS COCO dataset.  The GELAN and PGI-based object detection method surpassed all previous train-from-scratch methods in terms of object detection performance.  In terms of accuracy, the new method outperforms RT DETR~\cite{lv2023detrs} pre-trained with a large dataset, and it also outperforms depth-wise convolution-based design YOLO MS~\cite{chen2023yolo} in terms of parameters utilization.}	
		\label{fig:sota}	
		\vspace{-16pt}
	\end{figure}
	
	\section{Introduction}
	\label{sec:intr}
	
	\vspace{-4pt}
	
	Deep learning-based models have demonstrated far better performance than past artificial intelligence systems in various fields, such as computer vision, language processing, and speech recognition.  In recent years, researchers in the field of deep learning have mainly focused on how to develop more powerful system architectures and learning methods, such as CNNs~\cite{he2016deep,he2016identity,szegedy2016rethinking,huang2017densely,xie2017aggregated,liu2022convnext,woo2023convnext}, Transformers~\cite{dosovitskiy2021image,wang2021pyramid,wang2022pvt,liu2021swin,liu2022swin,ding2022davit,tu2022maxvit}, Perceivers~\cite{jaegle2021perceiver,jaegle2021perceiver,zhu2022uni,li2023uni,zhu2022uni,shridhar2023perceiver,tang2023perceiver}, and Mambas~\cite{gu2023mamba,zhu2024vision,liu2024vmamba}.  In addition, some researchers have tried to develop more general objective functions, such as loss function~\cite{zhou2019iou,rezatofighi2019generalized,chen2020ap, oksuz2020ranking,zheng2020distance,oksuz2021rank}, label assignment~\cite{zhu2020autoassign,ge2021ota,feng2021tood,wang2021end,li2022dual} and auxiliary supervision~\cite{lee2015deeply,szegedy2015going,wang2015training,shen2019object,levinshtein2020datnet,hayder2017boundary,huang2022monodtr,zhang2023monodetr,guo2020augfpn}.  The above studies all try to precisely find the mapping between input and target tasks.  However, most past approaches have ignored that input data may have a non-negligible amount of information loss during the feedforward process. This loss of information can lead to biased gradient flows, which are subsequently used to update the model. The above problems can result in deep networks to establish incorrect associations between targets and inputs, causing the trained model to produce incorrect predictions.

	\begin{figure*}[t]
		\begin{center}
			\includegraphics[width=1.\linewidth]{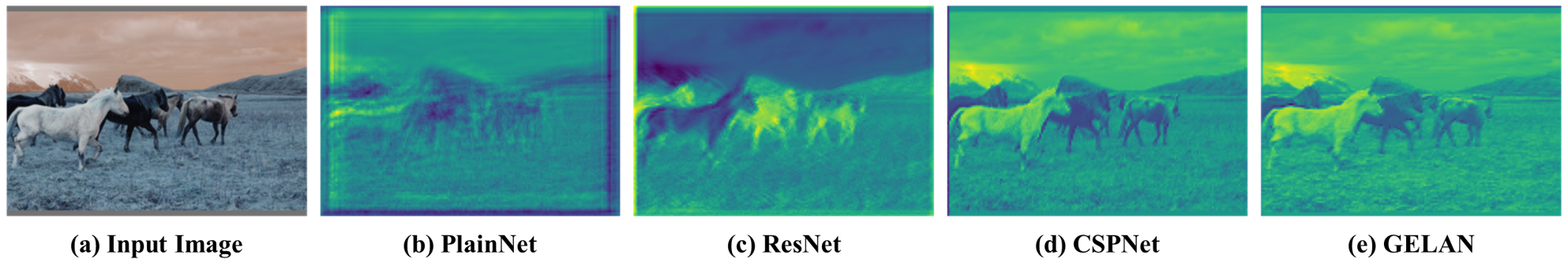}
		\end{center}
		\vspace{-14pt}
		\caption{Visualization results of random initial weight output feature maps for different network architectures: (a) input image, (b) PlainNet, (c) ResNet, (d) CSPNet, and (e) proposed GELAN.  From the figure, we can see that in different architectures, the information provided to the objective function to calculate the loss is lost to varying degrees, and our architecture can retain the most complete information and provide the most reliable gradient information for calculating the objective function.}	
		\vspace{-20pt}
		\label{fig:concept}
	\end{figure*}

	\newpage
	
	In deep networks, the phenomenon of input data losing information during the feedforward process is commonly known as information bottleneck~\cite{tishby2015deep}, and its schematic diagram is as shown in Figure~\ref{fig:concept}.  At present, the main methods that can alleviate this phenomenon are as follows: (1) The use of reversible architectures~\cite{gomez2017reversible,cai2022reversible,han2023revcolv2}: this method mainly uses repeated input data and maintains the information of the input data in an explicit way; (2) The use of masked modeling~\cite{kenton2019bert,xie2022simmim,chen2022sdae,bao2022beit,dosovitskiy2021image,woo2023convnext}: it mainly uses reconstruction loss and adopts an implicit way to maximize the extracted features and retain the input information; and (3) Introduction of the deep supervision concept~\cite{lee2015deeply,szegedy2015going,wang2015training,shen2019object}: it uses shallow features that have not lost too much important information to pre-establish a mapping from features to targets to ensure that important information can be transferred to deeper layers.  However, the above methods have different drawbacks in the training process and inference process. For example, a reversible architecture requires additional layers to combine repeatedly fed input data, which will significantly increase the inference cost.  In addition, since the input data layer to the output layer cannot have a too deep path, this limitation will make it difficult to model high-order semantic information during the training process.  As for masked modeling, its reconstruction loss sometimes conflicts with the target loss.  In addition, most mask mechanisms also produce incorrect associations with data.  For the deep supervision mechanism, it will produce error accumulation, and if the shallow supervision loses information during the training process, the subsequent layers will not be able to retrieve the required information.  The above phenomenon will be more significant on difficult tasks and small models.
	
	\vspace{-4pt}
	
	To address the above-mentioned issues, we propose a new concept, which is programmable gradient information (PGI).  The concept is to generate reliable gradients through auxiliary reversible branch, so that the deep features can still maintain key characteristics for executing target task.  The design of auxiliary reversible branch can avoid the semantic loss that may be caused by a traditional deep supervision process that integrates multi-path features. In other words, we are programming gradient information propagation at different semantic levels, and thereby achieving the best training results.  The reversible architecture of PGI is built on auxiliary branch, so there is no additional cost.  Since PGI can freely select loss function suitable for the target task, it also overcomes the problems encountered by mask modeling.  The proposed PGI mechanism can be applied to deep neural networks of various sizes and is more general than the deep supervision mechanism, which is only suitable for very deep neural networks.
	
	In this paper, we also designed generalized ELAN (GELAN) based on ELAN~\cite{wang2023designing}, the design of GELAN simultaneously takes into account the number of parameters, computational complexity, accuracy and inference speed.  This design allows users to arbitrarily choose appropriate computational blocks for different inference devices.  We combined the proposed PGI and GELAN, and then designed a new generation of YOLO series object detection system, which we call YOLOv9.  We used the MS COCO dataset to conduct experiments, and the experimental results verified that our proposed YOLOv9 achieved the top performance in all comparisons.
	
	We summarize the contributions of this paper as follows:
	\begin{enumerate}
		
		\item We theoretically analyzed the existing deep neural network architecture from the perspective of reversible function, and through this process we successfully explained many phenomena that were difficult to explain in the past.  We also designed PGI and auxiliary reversible branch based on this analysis and achieved excellent results.
		
		\vspace{-4pt}
		
		\item The PGI we designed solves the problem that deep supervision can only be used for extremely deep neural network architectures, and therefore allows new lightweight architectures to be truly applied in daily life.
		
		\vspace{-4pt}
		
		\item The GELAN we designed only uses conventional convolution to achieve a higher parameter usage than the depth-wise convolution design that based on the most advanced technology, while showing great advantages of being light, fast, and accurate.
		
		\vspace{-4pt}
		
		\item Combining the proposed PGI and GELAN, the object detection performance of the YOLOv9 on MS COCO dataset greatly surpasses the existing real-time object detectors in all aspects.
	\end{enumerate}	

	\newpage
	
	\section{Related work}
	\label{sec:relw}
	
	\subsection{Real-time Object Detectors}
	
	\vspace{-2pt}
	
	The current mainstream real-time object detectors are the YOLO series~\cite{redmon2016you,redmon2017yolo9000,redmon2018yolov3,bochkovskiy2020yolov4,wang2021scaled,ge2021yolox,xu2022pp,li2022yolov6,xu2022damo,glenn2022yolov5,li2023yolov6,huang2023yolocs,wang2023yolov7,chen2023yolo,wang2023gold,glenn2024yolov8}, and most of these models use CSPNet~\cite{wang2020cspnet} or ELAN~\cite{wang2023designing} and their variants as the main computing units.  In terms of feature integration, improved PAN~\cite{liu2018path} or FPN~\cite{lin2017feature} is often used as a tool, and then improved YOLOv3 head~\cite{redmon2018yolov3} or FCOS head~\cite{tian2019fcos,tian2022fcos} is used as prediction head.  Recently some real-time object detectors, such as RT DETR~\cite{lv2023detrs}, which puts its fundation on DETR~\cite{carion2020end}, have also been proposed.  However, since it is extremely difficult for DETR series object detector to be applied to new domains without a corresponding domain pre-trained model, the most widely used real-time object detector at present is still YOLO series.  This paper chooses YOLOv7~\cite{wang2023yolov7}, which has been proven effective in a variety of computer vision tasks and various scenarios, as a base to develop the proposed method.  We use GELAN to improve the architecture and the training process with the proposed PGI.  The above novel approach makes the proposed YOLOv9 the top real-time object detector of the new generation.
	
	\vspace{-2pt}
	
	\subsection{Reversible Architectures}
	
	\vspace{-2pt}
	
	The operation unit of reversible architectures~\cite{gomez2017reversible,cai2022reversible,han2023revcolv2} must maintain the characteristics of reversible conversion, so it can be ensured that the output feature map of each layer of operation unit can retain complete original information.  Before, RevCol~\cite{cai2022reversible} generalizes traditional reversible unit to multiple levels, and in doing so can expand the semantic levels expressed by different layer units. Through a literature review of various neural network architectures, we found that there are many high-performing architectures with varying degree of reversible properties.  For example, Res2Net module~\cite{gao2019res2net} combines different input partitions with the next partition in a hierarchical manner, and concatenates all converted partitions before passing them backwards.  CBNet~\cite{liu2020cbnet,liang2021cbnetv2} re-introduces the original input data through composite backbone to obtain complete original information, and obtains different levels of multi-level reversible information through various composition methods.  These network architectures generally have excellent parameter utilization, but the extra composite layers cause slow inference speeds.  DynamicDet~\cite{lin2023dynamicdet} combines CBNet~\cite{liang2021cbnetv2} and the high-efficiency real-time object detector YOLOv7~\cite{wang2023yolov7} to achieve a very good trade-off among speed, number of parameters, and accuracy.  This paper introduces the DynamicDet architecture as the basis for designing reversible branches.  In addition, reversible information is further introduced into the proposed PGI.  The proposed new architecture does not require additional connections during the inference process, so it can fully retain the advantages of speed, parameter amount, and accuracy.
	
	\subsection{Auxiliary Supervision}
	
	Deep supervision~\cite{lee2015deeply,szegedy2015going,wang2015training} is the most common auxiliary supervision method, which performs training by inserting additional prediction layers in the middle layers.  Especially the application of multi-layer decoders introduced in the transformer-based methods is the most common one.  Another common auxiliary supervision method is to utilize the relevant meta information to guide the feature maps produced by the intermediate layers and make them have the properties required by the target tasks~\cite{levinshtein2020datnet,hayder2017boundary,huang2022monodtr,zhang2023monodetr,guo2020augfpn}.  Examples of this type include using segmentation loss or depth loss to enhance the accuracy of object detectors.  Recently, there are many reports in the literature~\cite{wang2021end,sun2021makes,zong2023detrs} that use different label assignment methods to generate different auxiliary supervision mechanisms to speed up the convergence speed of the model and improve the robustness at the same time.  However, the auxiliary supervision mechanism is usually only applicable to large models, so when it is applied to lightweight models, it is easy to cause an under parameterization phenomenon, which makes the performance worse.  The PGI we proposed designed a way to reprogram multi-level semantic information, and this design allows lightweight models to also benefit from the auxiliary supervision mechanism. 
	
	\section{Problem Statement}
	
	Usually, people attribute the difficulty of deep neural network convergence problem due to factors such as gradient vanish or gradient saturation, and these phenomena do exist in traditional deep neural networks.  However, modern deep neural networks have already fundamentally solved the above problem by designing various normalization and activation functions.  Nevertheless, deep neural networks still have the problem of slow convergence or poor convergence results.  
	
	In this paper, we explore the nature of the above issue further.  Through in-depth analysis of information bottleneck, we deduced that the root cause of this problem is that the initial gradient originally coming from a very deep network has lost a lot of information needed to achieve the goal soon after it is transmitted. In order to confirm this inference, we feedforward deep networks of different architectures with initial weights, and then visualize and illustrate them in Figure~\ref{fig:concept}.  Obviously, PlainNet has lost a lot of important information required for object detection in deep layers.  As for the proportion of important information that ResNet, CSPNet, and GELAN can retain, it is indeed positively related to the accuracy that can be obtained after training.  We further design reversible network-based methods to solve the causes of the above problems.  In this section we shall elaborate our analysis of information bottleneck principle and reversible functions.

	\newpage
	
	\subsection{Information Bottleneck Principle}
	
	According to information bottleneck principle, we know that data $X$ may cause information loss when going through transformation, as shown in Eq.~\ref{eq:ibp} below:
	
	\begin{equation}
		I(X,X) \geq I(X, f_{\theta}(X)) \geq I(X, g_{\phi}(f_{\theta}(X))),
		\label{eq:ibp}
	\end{equation}
	where $I$ indicates mutual information, $f$ and $g$ are transformation functions, and $\theta$ and $\phi$ are parameters of $f$ and $g$, respectively.
	
	In deep neural networks, $f_{\theta}(\cdot)$ and $g_{\phi}(\cdot)$ respectively represent the operations of two consecutive layers in deep neural network.  From Eq.~\ref{eq:ibp}, we can predict that as the number of network layer becomes deeper, the original data will be more likely to be lost.  However, the parameters of the deep neural network are based on the output of the network as well as the given target, and then update the network after generating new gradients by calculating the loss function.  As one can imagine, the output of a deeper neural network is less able to retain complete information about the prediction target.  This will make it possible to use incomplete information during network training, resulting in unreliable gradients and poor convergence.
	
	One way to solve the above problem is to directly increase the size of the model.  When we use a large number of parameters to construct a model, it is more capable of performing a more complete transformation of the data.  The above approach allows even if information is lost during the data feedforward process, there is still a chance to retain enough information to perform the mapping to the target.  The above phenomenon explains why the width is more important than the depth in most modern models.  However, the above conclusion cannot fundamentally solve the problem of unreliable gradients in very deep neural network.  Below, we will introduce how to use reversible functions to solve problems and conduct relative analysis.
	
	\subsection{Reversible Functions}
	
	When a function $r$ has an inverse transformation function $v$, we call this function reversible function, as shown in Eq.~\ref{eq:rf}.
	
	\begin{equation}
	X = v_{\zeta}(r_{\psi}(X)),
	\label{eq:rf}
	\end{equation}
	where $\psi$ and $\zeta$ are parameters of $r$ and $v$, respectively.  Data $X$ is converted by reversible function without losing information, as shown in Eq.~\ref{eq:ll}.
	
	\begin{equation}
	I(X,X) = I(X, r_{\psi}(X)) = I(X, v_{\zeta}(r_{\psi}(X))).
	\label{eq:ll}
	\end{equation}		
	When the network's transformation function is composed of reversible functions, more reliable gradients can be obtained to update the model.  Almost all of today's popular deep learning methods are architectures that conform to the reversible property, such as Eq.~\ref{eq:resnet}.
	
	\begin{equation}
	X^{l+1} = X^{l} + f^{l+1}_{\theta}(X^{l}),
	\label{eq:resnet}
	\end{equation}
	where $l$ indicates the $l$-th layer of a PreAct ResNet and $f$ is the transformation function of the $l$-th layer.  PreAct ResNet~\cite{he2016identity} repeatedly passes the original data $X$ to subsequent layers in an explicit way.  Although such a design can make a deep neural network with more than a thousand layers converge very well, it destroys an important reason why we need deep neural networks. That is, for difficult problems, it is difficult for us to directly find simple mapping functions to map data to targets.  This also explains why PreAct ResNet performs worse than ResNet~\cite{he2016deep} when the number of layers is small.
	
	In addition, we tried to use masked modeling that allowed the transformer model to achieve significant breakthroughs.  We use approximation methods, such as Eq.~\ref{eq:mm}, to try to find the inverse transformation $v$ of $r$, so that the transformed features can retain enough information using sparse features.  The form of Eq.~\ref{eq:mm} is as follows:
	
	\begin{equation}
	X = v_{\zeta}(r_{\psi}(X) \cdot M),
	\label{eq:mm}
	\end{equation}
	where $M$ is a dynamic binary mask.  Other methods that are commonly used to perform the above tasks are diffusion model and variational autoencoder, and they both have the function of finding the inverse function.  However, when we apply the above approach to a lightweight model, there will be defects because the lightweight model will be under parameterized to a large amount of raw data.  Because of the above reason, important information $I(Y,X)$ that maps data $X$ to target $Y$ will also face the same problem.  For this issue, we will explore it using the concept of information bottleneck~\cite{tishby2015deep}. The formula for information bottleneck is as follows:
	
	\begin{equation}
	I(X,X) \geq I(Y,X) \geq I(Y, f_{\theta}(X)) \geq ... \geq I(Y,\hat{Y}).
	\label{eq:ibpy}
	\end{equation}		
	Generally speaking, $I(Y,X)$ will only occupy a very small part of $I(X,X)$.  However, it is critical to the target mission.  Therefore, even if the amount of information lost in the feedforward stage is not significant, as long as $I(Y,X)$ is covered, the training effect will be greatly affected.  The lightweight model itself is in an under parameterized state, so it is easy to lose a lot of important information in the feedforward stage.  Therefore, our goal for the lightweight model is how to accurately filter $I(Y,X)$ from $I(X,X)$.  As for fully preserving the information of $X$, that is difficult to achieve.  Based on the above analysis, we hope to propose a new deep neural network training method that can not only generate reliable gradients to update the model, but also be suitable for shallow and lightweight neural networks.
	
	\begin{figure*}[t]
		\begin{center}
			\includegraphics[width=1.\linewidth]{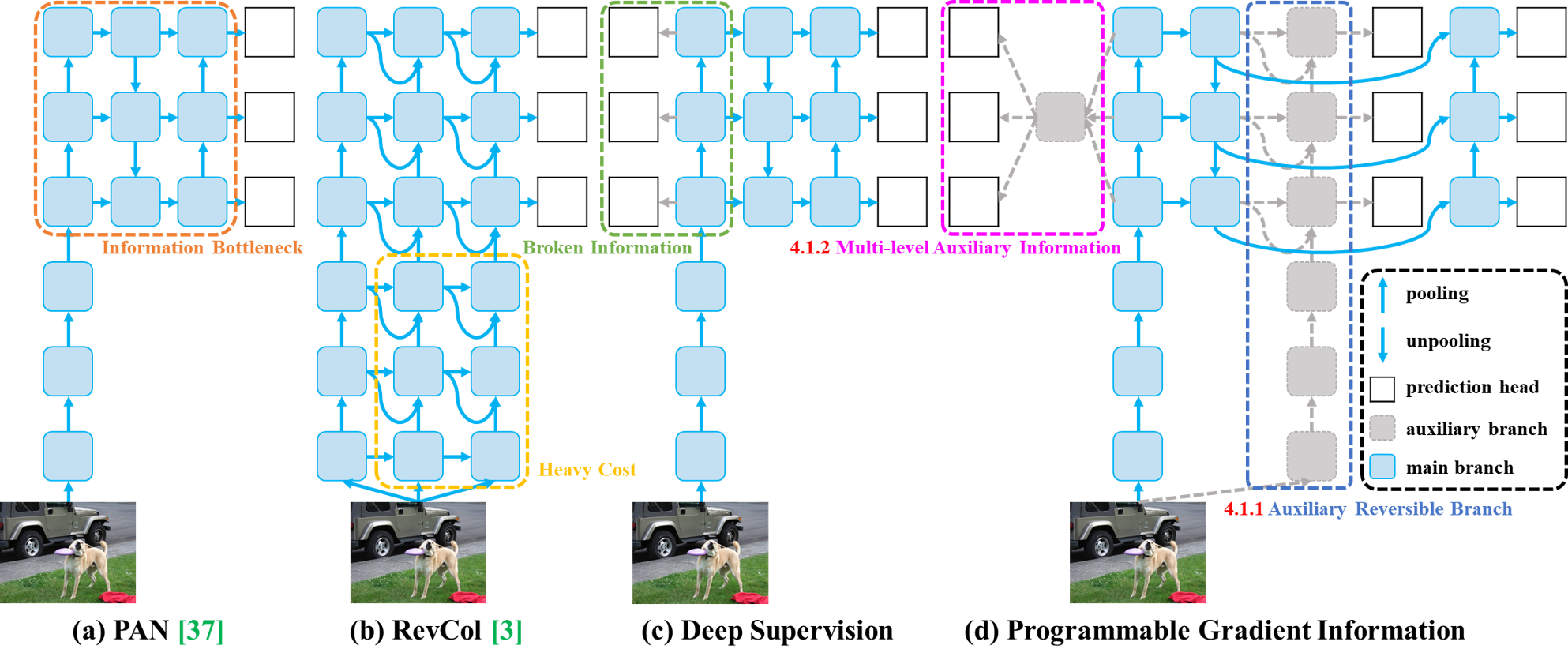}
		\end{center}
		\vspace{-12pt}
		\caption{PGI and related network architectures and methods.  (a) Path Aggregation Network (PAN))~\cite{liu2018path}, (b) Reversible Columns (RevCol)~\cite{cai2022reversible}, (c) conventional deep supervision, and (d) our proposed Programmable Gradient Information (PGI).  PGI is mainly composed of three components: (1) main branch: architecture used for inference, (2) auxiliary reversible branch: generate reliable gradients to supply main branch for backward transmission, and (3) multi-level auxiliary information: control main branch learning plannable multi-level of semantic information.}		
		\vspace{-12pt}
		\label{fig:pgi}
	\end{figure*}

	\newpage
	
	\section{Methodology}
	
	\subsection{Programmable Gradient Information}
	
	In order to solve the aforementioned problems, we propose a new auxiliary supervision framework called Programmable Gradient Information (PGI), as shown in Figure~\ref{fig:pgi} (d).  PGI mainly includes three components, namely (1) main branch, (2) auxiliary reversible branch, and (3) multi-level auxiliary information.  From Figure~\ref{fig:pgi} (d) we see that the inference process of PGI only uses main branch and therefore does not require any additional inference cost.  As for the other two components, they are used to solve or slow down several important issues in deep learning methods.  Among them, auxiliary reversible branch is designed to deal with the problems caused by the deepening of neural networks. Network deepening will cause information bottleneck, which will make the loss function unable to generate reliable gradients.  As for multi-level auxiliary information, it is designed to handle the error accumulation problem caused by deep supervision, especially for the architecture and lightweight model of multiple prediction branch.  Next, we will introduce these two components step by step.
	
	\subsubsection{Auxiliary Reversible Branch}
	
	In PGI, we propose auxiliary reversible branch to generate reliable gradients and update network parameters.  By providing information that maps from data to targets, the loss function can provide guidance and avoid the possibility of finding false correlations from incomplete feedforward features that are less relevant to the target.  We propose the maintenance of complete information by introducing reversible architecture, but adding main branch to reversible architecture will consume a lot of inference costs.  We analyzed the architecture of Figure~\ref{fig:pgi} (b) and found that when additional connections from deep to shallow layers are added, the inference time will increase by 20\%.  When we repeatedly add the input data to the high-resolution computing layer of the network (yellow box), the inference time even exceeds twice the time.
	
	Since our goal is to use reversible architecture to obtain reliable gradients, “reversible” is not the only necessary condition in the inference stage.  In view of this, we regard reversible branch as an expansion of deep supervision branch, and then design auxiliary reversible branch, as shown in Figure~\ref{fig:pgi} (d).  As for the main branch deep features that would have lost important information due to information bottleneck, they will be able to receive reliable gradient information from the auxiliary reversible branch. These gradient information will drive parameter learning to assist in extracting correct and important information, and the above actions can enable the main branch to obtain features that are more effective for the target task.  Moreover, the reversible architecture performs worse on shallow networks than on general networks because complex tasks require conversion in deeper networks.  Our proposed method does not force the main branch to retain complete original information but updates it by generating useful gradient through the auxiliary supervision mechanism.  The advantage of this design is that the proposed method can also be applied to shallower networks.
	
	\begin{figure*}[t]
		\begin{center}
			\includegraphics[width=.8\linewidth]{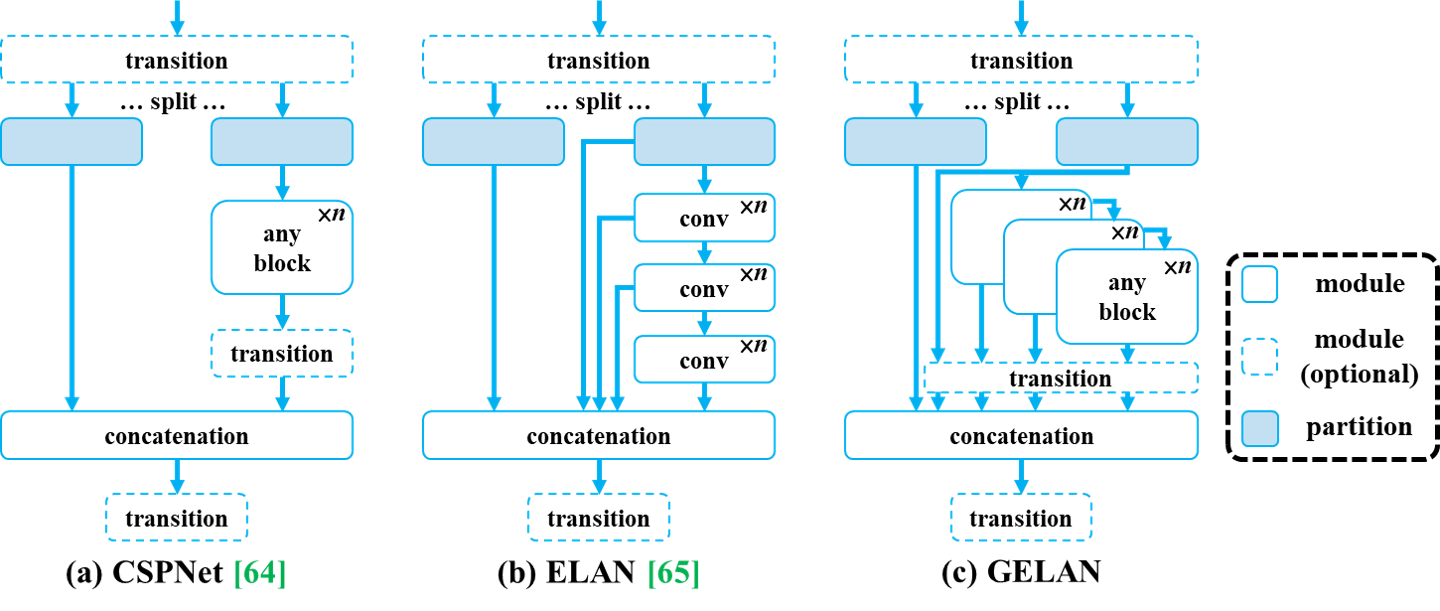}
		\end{center}
		\vspace{-14pt}
		\caption{The architecture of GELAN: (a) CSPNet~\cite{wang2020cspnet}, (b) ELAN~\cite{wang2023designing}, and (c) proposed GELAN.  We imitate CSPNet and extend ELAN into GELAN that can support any computational blocks.}	
		\vspace{-18pt}
		\label{fig:gelan}
	\end{figure*}

	\newpage
	
	Finally, since auxiliary reversible branch can be removed during the inference phase, the inference capabilities of the original network can be retained. We can also choose any reversible architectures in PGI to play the role of auxiliary reversible branch.
	
	\vspace{-12pt}
	
	\subsubsection{Multi-level Auxiliary Information}
	
	In this section we will discuss how multi-level auxiliary information works.  The deep supervision architecture including multiple prediction branch is shown in Figure~\ref{fig:pgi} (c).  For object detection, different feature pyramids can be used to perform different tasks, for example together they can detect objects of different sizes.  Therefore, after connecting to the deep supervision branch, the shallow features will be guided to learn the features required for small object detection, and at this time the system will regard the positions of objects of other sizes as the background.  However, the above deed will cause the deep feature pyramids to lose a lot of information needed to predict the target object. Regarding this issue, we believe that each feature pyramid needs to receive information about all target objects so that subsequent main branch can retain complete information to learn predictions for various targets.	
	
	The concept of multi-level auxiliary information is to insert an integration network between the feature pyramid hierarchy layers of auxiliary supervision and the main branch, and then uses it to combine returned gradients from different prediction heads, as shown in Figure~\ref{fig:pgi} (d).  Multi-level auxiliary information is then to aggregate the gradient information containing all target objects, and pass it to the main branch and then update parameters.  At this time, the characteristics of the main branch's feature pyramid hierarchy will not be dominated by some specific object's information.  As a result, our method can alleviate the broken information problem in deep supervision.  In addition, any integrated network can be used in multi-level auxiliary information.  Therefore, we can plan the required semantic levels to guide the learning of network architectures of different sizes. 
	
	\subsection{Generalized ELAN}
	
	In this Section we describe the proposed new network architecture -- GELAN. By combining two neural network architectures, CSPNet~\cite{wang2020cspnet} and ELAN~\cite{wang2023designing}, which are designed with gradient path planning, we designed generalized efficient layer aggregation network (GELAN) that takes into account lighweight, inference speed, and accuracy. Its overall architecture is shown in Figure~\ref{fig:gelan}. We generalized the capability of ELAN~\cite{wang2023designing}, which originally only used stacking of convolutional layers, to a new architecture that can use any computational blocks.
	
	\begin{table*}[t]
		\centering
		\begin{threeparttable}[t]
			\footnotesize
			\caption{Comparison of state-of-the-art real-time object detectors.}
			\label{table:sota}
			\setlength\tabcolsep{4.0pt}
			\begin{tabular}{lcccccccc}
				\toprule
				\textbf{Model} & \textbf{\#Param. (M)} & \textbf{FLOPs (G)} & \textbf{AP$^{val}_{50:95}$ (\%)} & \textbf{AP$^{val}_{50}$ (\%)} & \textbf{AP$^{val}_{75}$ (\%)} & \textbf{AP$^{val}_{S}$ (\%)} & \textbf{AP$^{val}_{M}$ (\%)} & \textbf{AP$^{val}_{L}$ (\%)} \\
				\midrule
				\textbf{YOLOv5-N r7.0~\cite{glenn2022yolov5}} & 1.9 & 4.5 & 28.0 & 45.7 & -- & -- & -- & -- \\
				\textbf{YOLOv5-S r7.0~\cite{glenn2022yolov5}} & 7.2 & 16.5 & 37.4 & 56.8 & -- & -- & -- & -- \\
				\textbf{YOLOv5-M r7.0~\cite{glenn2022yolov5}} & 21.2 & 49.0 & 45.4 & 64.1 & -- & -- & -- & -- \\
				\textbf{YOLOv5-L r7.0~\cite{glenn2022yolov5}} & 46.5 & 109.1 & 49.0 & 67.3 & -- & -- & -- & -- \\
				\textbf{YOLOv5-X r7.0~\cite{glenn2022yolov5}} & 86.7 & 205.7 & 50.7 & 68.9 & -- & -- & -- & -- \\
				\midrule
				\textbf{YOLOv6-N v3.0~\cite{li2023yolov6}} & 4.7 & 11.4 & 37.0 & 52.7 & -- & -- & -- & -- \\
				\textbf{YOLOv6-S v3.0~\cite{li2023yolov6}} & 18.5 & 45.3 & 44.3 & 61.2 & -- & -- & -- & -- \\
				\textbf{YOLOv6-M v3.0~\cite{li2023yolov6}} & 34.9 & 85.8 & 49.1 & 66.1 & -- & -- & -- & -- \\
				\textbf{YOLOv6-L v3.0~\cite{li2023yolov6}} & 59.6 & 150.7 & 51.8 & 69.2 & -- & -- & -- & -- \\
				\midrule
				\textbf{YOLOv7~\cite{wang2023yolov7}} & 36.9 & 104.7 & 51.2 & 69.7 & 55.9 & 31.8 & 55.5 & 65.0 \\
				\textbf{YOLOv7-X~\cite{wang2023yolov7}} & 71.3 & 189.9 & 52.9 & 71.1 & 51.4 & 36.9 & 57.7 & 68.6 \\
				\midrule
				\textbf{YOLOv7-N AF~\cite{wang2023yolov7}} & 3.1 & 8.7 & 37.6 & 53.3 & 40.6 & 18.7 & 41.7 & 52.8 \\
				\textbf{YOLOv7-S AF~\cite{wang2023yolov7}} & 11.0 & 28.1 & 45.1 & 61.8 & 48.9 & 25.7 & 50.2 & 61.2 \\
				\textbf{YOLOv7 AF~\cite{wang2023yolov7}} & 43.6 & 130.5 & 53.0 & 70.2 & 57.5 & 35.8 & 58.7 & 68.9 \\
				\midrule
				\textbf{YOLOv8-N~\cite{glenn2024yolov8}} & 3.2 & 8.7 & 37.3 & 52.6 & -- & -- & -- & -- \\
				\textbf{YOLOv8-S~\cite{glenn2024yolov8}} & 11.2 & 28.6 & 44.9 & 61.8 & -- & -- & -- & -- \\
				\textbf{YOLOv8-M~\cite{glenn2024yolov8}} & 25.9 & 78.9 & 50.2 & 67.2 & -- & -- & -- & -- \\
				\textbf{YOLOv8-L~\cite{glenn2024yolov8}} & 43.7 & 165.2 & 52.9 & 69.8 & 57.5 & 35.3 & 58.3 & 69.8 \\
				\textbf{YOLOv8-X~\cite{glenn2024yolov8}} & 68.2 & 257.8 & 53.9 & 71.0 & 58.7 & 35.7 & 59.3 & 70.7 \\
				\midrule
				\textbf{DAMO YOLO-T~\cite{xu2022damo}} & 8.5 & 18.1 & 42.0 & 58.0 & 45.2 & 23.0 & 46.1 & 58.5 \\
				\textbf{DAMO YOLO-S~\cite{xu2022damo}} & 12.3 & 37.8 & 46.0 & 61.9 & 49.5 & 25.9 & 50.6 & 62.5 \\
				\textbf{DAMO YOLO-M~\cite{xu2022damo}} & 28.2 & 61.8 & 49.2 & 65.5 & 53.0 & 29.7 & 53.1 & 66.1 \\
				\textbf{DAMO YOLO-L~\cite{xu2022damo}} & 42.1 & 97.3 & 50.8 & 67.5 & 55.5 & 33.2 & 55.7 & 66.6 \\
				\midrule
				\textbf{Gold YOLO-N~\cite{wang2023gold}} & 5.6 & 12.1 & 39.6 & 55.7 & -- & 19.7 & 44.1 & 57.0 \\
				\textbf{Gold YOLO-S~\cite{wang2023gold}} & 21.5 & 46.0 & 45.4 & 62.5 & -- & 25.3 & 50.2 & 62.6 \\
				\textbf{Gold YOLO-M~\cite{wang2023gold}} & 41.3 & 87.5 & 49.8 & 67.0 & -- & 32.3 & 55.3 & 66.3 \\
				\textbf{Gold YOLO-L~\cite{wang2023gold}} & 75.1 & 151.7 & 51.8 & 68.9 & -- & 34.1 & 57.4 & 68.2 \\
				\midrule
				\textbf{YOLO MS-N~\cite{chen2023yolo}} & 4.5 & 17.4 & 43.4 & 60.4 & 47.6 & 23.7 & 48.3 & 60.3 \\
				\textbf{YOLO MS-S~\cite{chen2023yolo}} & 8.1 & 31.2 & 46.2 & 63.7 & 50.5 & 26.9 & 50.5 & 63.0 \\
				\textbf{YOLO MS~\cite{chen2023yolo}} & 22.2 & 80.2 & 51.0 & 68.6 & 55.7 & 33.1 & 56.1 & 66.5 \\
				\midrule
				\textbf{GELAN-S (Ours)} & 7.1 & 26.4 & 46.7 & 63.0 & 50.7 & 25.9 & 51.5 & 64.0 \\
				\textbf{GELAN-M (Ours)} & 20.0 & 76.3 & 51.1 & 67.9 & 55.7 & 33.6 & 56.4 & 67.3 \\
				\textbf{GELAN-C (Ours)} & 25.3 & 102.1 & 52.5 & 69.5 & 57.3 & 35.8 & 57.6 & 69.4 \\
				\textbf{GELAN-E (Ours)} & 57.3 & 189.0 & 55.0 & 71.9 & 60.0 & 38.0 & 60.6 & 70.9 \\
				\midrule
				\textbf{YOLOv9-S (Ours)} & 7.1 & 26.4 & 46.8 & 63.4 & 50.7 & 26.6 & 56.0 & 64.5 \\
				\textbf{YOLOv9-M (Ours)} & 20.0 & 76.3 & 51.4 & 68.1 & 56.1 & 33.6 & 57.0 & 68.0 \\
				\textbf{YOLOv9-C (Ours)} & 25.3 & 102.1 & 53.0 & 70.2 & 57.8 & 36.2 & 58.5 & 69.3 \\
				\textbf{YOLOv9-E (Ours)} & 57.3 & 189.0 & 55.6 & 72.8 & 60.6 & 40.2 & 61.0 & 71.4 \\
				\bottomrule
			\end{tabular}
		\end{threeparttable}
		\vspace{-8pt}
	\end{table*}
	
	\section{Experiments}
	
	\subsection{Experimental Setup}
	
	We verify the proposed method with MS COCO dataset. All experimental setups follow YOLOv7 AF~\cite{wang2023yolov7}, while the dataset is MS COCO 2017 splitting. All models we mentioned are trained using the train-from-scratch strategy, and the total number of training times is 500 epochs. In setting the learning rate, we use linear warm-up in the first three epochs, and the subsequent epochs set the corresponding decay manner according to the model scale. As for the last 15 epochs, we turn mosaic data augmentation off. For more settings, please refer to Appendix.
	
	\subsection{Implimentation Details}
	
	We built general and extended version of YOLOv9 based on YOLOv7~\cite{wang2023yolov7} and Dynamic YOLOv7~\cite{lin2023dynamicdet} respectively. In the design of the network architecture, we replaced ELAN~\cite{wang2023designing} with GELAN using CSPNet blocks~\cite{wang2020cspnet} with planned RepConv~\cite{wang2023yolov7} as computational blocks. We also simplified downsampling module and optimized anchor-free prediction head. As for the auxiliary loss part of PGI, we completely follow YOLOv7's auxiliary head setting. Please see Appendix for more details.
	
	\newpage
	
	\subsection{Comparison with state-of-the-arts}
	\label{sec:cmp}
	
	\vspace{-6pt}
	
	Table~\ref{table:sota} lists comparison of our proposed YOLOv9 with other train-from-scratch real-time object detectors. Overall, the best performing methods among existing methods are YOLO MS-S~\cite{chen2023yolo} for lightweight models, YOLO MS~\cite{chen2023yolo} for medium models, YOLOv7 AF~\cite{wang2023yolov7} for general models, and YOLOv8-X~\cite{glenn2024yolov8} for large models. Compared with lightweight and medium model YOLO MS~\cite{chen2023yolo}, YOLOv9 has about 10\% less parameters and 5$\sim$15\% less calculations, but still has a 0.4$\sim$0.6\% improvement in AP. Compared with YOLOv7 AF, YOLOv9-C has 42\% less parameters and 22\% less calculations, but achieves the same AP (53\%). Compared with YOLOv8-X, YOLOv9-E has 16\% less parameters, 27\% less calculations, and has significant improvement of 1.7\% AP. The above comparison results show that our proposed YOLOv9 has significantly improved in all aspects compared with existing methods.
	
	\begin{figure*}[t]
		\begin{center}
			\includegraphics[width=.85\linewidth]{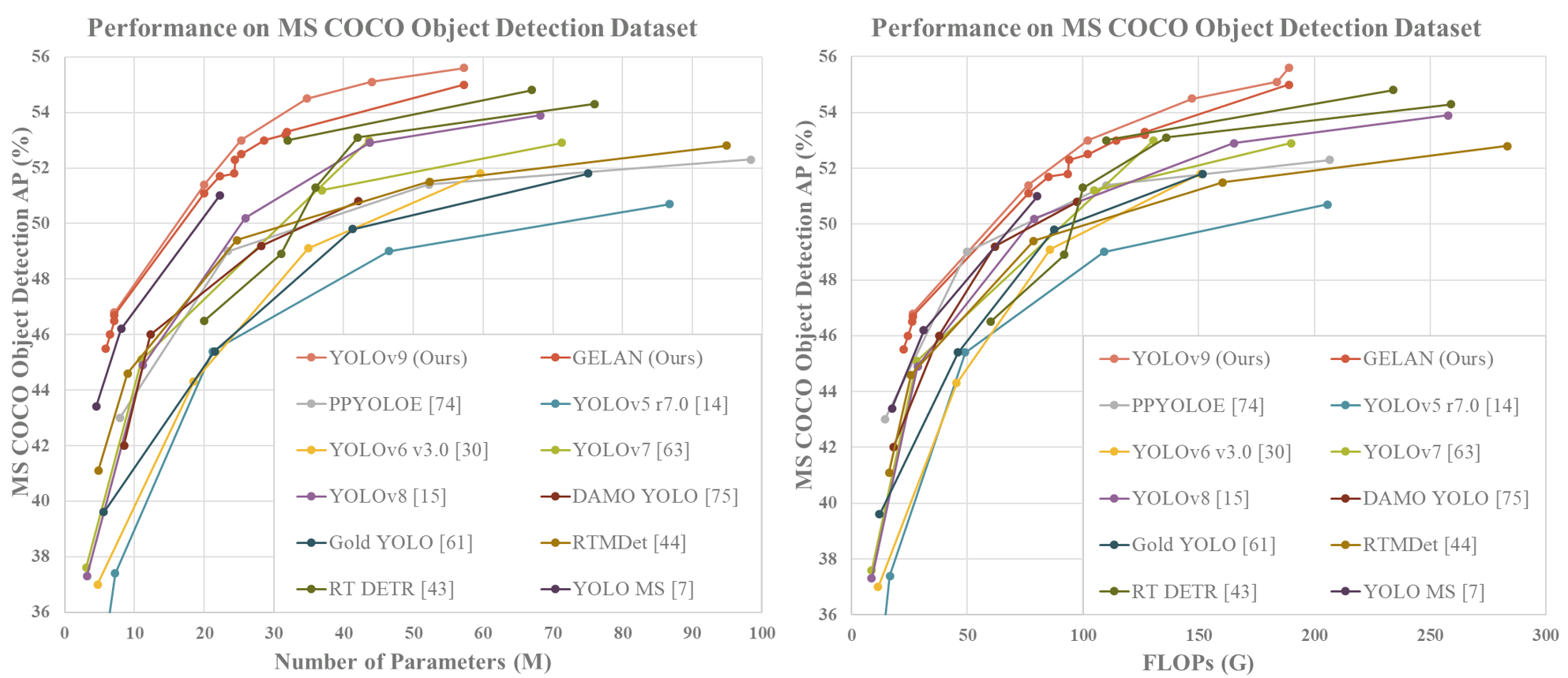}
		\end{center}
		\vspace{-6pt}
		\caption{Comparison of state-of-the-art real-time object detectors. The methods participating in the comparison all use ImageNet as pre-trained weights, including RT DETR~\cite{lv2023detrs}, RTMDet~\cite{lyu2022rtmdet}, and PP-YOLOE~\cite{xu2022pp}, etc.  The YOLOv9 that uses train-from-scratch method clearly surpasses the performance of other methods.}	
		\vspace{-12pt}
		\label{fig:perf}
	\end{figure*}
	
	On the other hand, we also include ImageNet pretrained model in the comparison, and the results are shown in Figure~\ref{fig:perf}.  We compare them based on the parameters and the amount of computation respectively.  In terms of the number of parameters, the best performing large model is RT DETR~\cite{lv2023detrs}.  From Figure~\ref{fig:perf}, we can see that YOLOv9 using conventional convolution is even better than YOLO MS using depth-wise convolution in parameter utilization.  As for the parameter utilization of large models, it also greatly surpasses RT DETR using ImageNet pretrained model.  Even better is that in the deep model, YOLOv9 shows the huge advantages of using PGI.  By accurately retaining and extracting the information needed to map the data to the target, our method requires only 66\% of the parameters while maintaining the accuracy as RT DETR-X.
	
	\newpage
	
	As for the amount of computation, the best existing models from the smallest to the largest are YOLO MS~\cite{chen2023yolo}, PP YOLOE~\cite{xu2022pp}, and RT DETR~\cite{lv2023detrs}.  From Figure 5, we can see that YOLOv9 is far superior to the train-from-scratch methods in terms of computational complexity.  In addition, if compared with those based on depth-wise convolution and ImageNet-based pretrained models, YOLOv9 is also very competitive.
	
	\subsection{Ablation Studies}
	\label{sec:abl}
	
	\subsubsection{Generalized ELAN}
	
	For GELAN, we first do ablation studies for computational blocks.  We used Res blocks~\cite{he2016deep}, Dark blocks~\cite{redmon2018yolov3}, and CSP blocks~\cite{wang2020cspnet} to conduct experiments, respectively.  Table~\ref{table:cb} shows that after replacing convolutional layers in ELAN with different computational blocks, the system can maintain good performance.  Users are indeed free to replace computational blocks and use them on their respective inference devices.  Among different computational block replacements, CSP blocks perform particularly well.  They not only reduce the amount of parameters and computation, but also improve AP by 0.7\%.  Therefore, we choose CSP-ELAN as the component unit of GELAN in YOLOv9.
	
	\begin{table}[h]
		\centering
		\begin{threeparttable}[h]
			\footnotesize
			\caption{Ablation study on various computational blocks.}
			\label{table:cb}
			\begin{tabular}{lcccc}
				\toprule
				\textbf{Model} & \textbf{CB type} & \textbf{\#Param.} & \textbf{FLOPs} & \textbf{AP$^{val}_{50:95}$} \\	
				\midrule
				\textbf{GELAN-S} & Conv & 6.2M & 23.5G & 44.8\% \\
				\textbf{GELAN-S} & Res~\cite{he2016deep} & 5.4M & 21.0G & 44.3\% \\
				\textbf{GELAN-S} & Dark~\cite{redmon2018yolov3} & 5.7M & 21.8G & 44.5\% \\
				\textbf{GELAN-S} & CSP~\cite{wang2020cspnet} & 5.9M & 22.4G & 45.5\% \\
				\bottomrule
			\end{tabular}
			\begin{tablenotes}[flushleft]
			\footnotesize
			\item[1] CB type nedotes as computational block type.
			\item[2] -S nedotes small size model.
			\end{tablenotes}
		\end{threeparttable}
	\end{table}

	\newpage
	
	Next, we conduct ELAN block-depth and CSP block-depth experiments on GELAN of different sizes, and display the results in Table~\ref{table:depth}.  We can see that when the depth of ELAN is increased from 1 to 2, the accuracy is significantly improved.  But when the depth is greater than or equal to 2, no matter it is improving the ELAN depth or the CSP depth, the number of parameters, the amount of computation, and the accuracy will always show a linear relationship.  This means GELAN is not sensitive to the depth.  In other words, users can arbitrarily combine the components in GELAN to design the network architecture, and have a model with stable performance without special design.  In Table 3, for YOLOv9-\{S,M,C\}, we set the pairing of the ELAN depth and the CSP depth to \{\{2, 3\}, \{2, 1\}, \{2, 1\}\}.

	\begin{table}[h]
	\centering
	\begin{threeparttable}[h]
		\footnotesize
		\caption{Ablation study on ELAN and CSP depth.}
		\label{table:depth}
		\setlength\tabcolsep{5.0pt}
		\begin{tabular}{lccccc}
			\toprule
			\textbf{Model} & \textbf{D$_{ELAN}$} & \textbf{D$_{CSP}$} & \textbf{\#Param.} & \textbf{FLOPs} & \textbf{AP$^{val}_{50:95}$} \\	
			\midrule
			\textbf{GELAN-S} & 2 & 1 & 5.9M & 22.4G & 45.5\% \\
			\textbf{GELAN-S} & 2 & 2 & 6.5M & 24.4G & 46.0\% \\
			\textbf{GELAN-S} & 3 & 1 & 7.1M & 26.3G & 46.5\% \\
			\textbf{GELAN-S} & 2 & 3 & 7.1M & 26.4G & 46.7\% \\
			\midrule
			\textbf{GELAN-M} & 2 & 1 & 20.0M & 76.3G & 51.1\% \\
			\textbf{GELAN-M} & 2 & 2 & 22.2M & 85.1G & 51.7\% \\
			\textbf{GELAN-M} & 3 & 1 & 24.3M & 93.5G & 51.8\% \\
			\textbf{GELAN-M} & 2 & 3 & 24.4M & 94.0G & 52.3\% \\
			\midrule
			\textbf{GELAN-C} & 1 & 1 & 18.9M & 77.5G & 50.7\% \\
			\textbf{GELAN-C} & 2 & 1 & 25.3M & 102.1G & 52.5\% \\
			\textbf{GELAN-C} & 2 & 2 & 28.6M & 114.4G & 53.0\% \\
			\textbf{GELAN-C} & 3 & 1 & 31.7M & 126.8G & 53.2\% \\
			\textbf{GELAN-C} & 2 & 3 & 31.9M & 126.7G & 53.3\% \\
			\bottomrule
		\end{tabular}
		\begin{tablenotes}[flushleft]
			\footnotesize
			\item[1] \textbf{D$_{ELAN}$} and \textbf{D$_{CSP}$} respectively nedotes depth of ELAN and CSP.
			\item[2] -\{S, M, C\} indicate small, medium, and compact models.
		\end{tablenotes}
	\end{threeparttable}
	\end{table}

	\newpage
	
	\subsubsection{Programmable Gradient Information}
	
	In terms of PGI, we performed ablation studies on auxiliary reversible branch and multi-level auxiliary information on the backbone and neck, respectively.  We designed auxiliary reversible branch ICN to use DHLC~\cite{liang2021cbnetv2} linkage to obtain multi-level reversible information.  As for multi-level auxiliary information, we use FPN and PAN for ablation studies and the role of PFH is equivalent to the traditional deep supervision.  The results of all experiments are listed in Table~\ref{table:pgi}.  From Table~\ref{table:pgi}, we can see that PFH is only effective in deep models, while our proposed PGI can improve accuracy under different combinations.  Especially when using ICN, we get stable and better results.  We also tried to apply the lead-head guided assignment proposed in YOLOv7~\cite{wang2023yolov7} to the PGI's auxiliary supervision, and achieved much better performance.
	
	\begin{table}[h]
		\centering
		\begin{threeparttable}[h]
			\footnotesize
			\caption{Ablation study on PGI of backbone and neck.}
			\label{table:pgi}
			\setlength\tabcolsep{3.0pt}
			\begin{tabular}{lcccccc}
				\toprule
				\textbf{Model} & \textbf{G$_{backbone}$} & \textbf{G$_{neck}$} & \textbf{AP$^{val}_{50:95}$} & \textbf{AP$^{val}_{S}$} & \textbf{AP$^{val}_{M}$} & \textbf{AP$^{val}_{L}$} \\
				\textbf{GELAN-C} & -- & -- & 52.5\% & 35.8\% & 57.6\% & \textbf{69.4\%} \\
				\textbf{GELAN-C} & PFH & -- & 52.5\% & 35.3\% & 58.1\% & 68.9\% \\
				\textbf{GELAN-C} & FPN & -- & 52.6\% & 35.3\% & 58.1\% & 68.9\% \\
				\textbf{GELAN-C} & -- & ICN & 52.7\% & 35.3\% & 58.4\% & 68.9\% \\
				\textbf{GELAN-C} & FPN & ICN & 52.8\% & 35.8\% & 58.2\% & 69.1\% \\	
				\textbf{GELAN-C} & ICN & -- & \textbf{52.9\%} & 35.2\% & \textbf{58.7\%} & 68.6\% \\
				\textbf{GELAN-C} & LHG-ICN & -- & \textbf{53.0\%} & \textbf{36.3\%} & 58.5\% & 69.1\% \\
				\midrule
				\textbf{GELAN-E} & -- & -- & 55.0\% & 38.0\% & 60.6\% & 70.9\% \\
				\textbf{GELAN-E} & PFH & -- & 55.3\% & 38.3\% & 60.3\% & 71.6\% \\
				\textbf{GELAN-E} & FPN & -- & \textbf{55.6\%} & \textbf{40.2\%} & 61.0\% & 71.4\% \\
				\textbf{GELAN-E} & PAN & -- & 55.5\% & 39.0\% & \textbf{61.1\%} & 71.5\% \\
				\textbf{GELAN-E} & FPN & ICN & \textbf{55.6\%} & 39.8\% & 60.9\% & \textbf{71.9\%} \\
				\bottomrule
			\end{tabular}
			\begin{tablenotes}[flushleft]
				\footnotesize
				\item[1] \textbf{D$_{ELAN}$} and \textbf{D$_{CSP}$} respectively nedotes depth of ELAN and CSP.
				\item[2] LHG indicates lead head guided training proposed by YOLOv7~\cite{wang2023yolov7}.
			\end{tablenotes}
		\end{threeparttable}
	\end{table}
	
	We further implemented the concepts of PGI and deep supervision on models of various sizes and compared the results, these results are shown in Table~\ref{table:scale}.  As analyzed at the beginning, introduction of deep supervision will cause a loss of accuracy for shallow models.  As for general models, introducing deep supervision will cause unstable performance, and the design concept of deep supervision can only bring gains in extremely deep models.  The proposed PGI can effectively handle problems such as information bottleneck and information broken, and can comprehensively improve the accuracy of models of different sizes.  The concept of PGI brings two valuable contributions.  The first one is to make the auxiliary supervision method applicable to shallow models, while the second one is to make the deep model training process obtain more reliable gradients.  These gradients enable deep models to use more accurate information to establish correct correlations between data and targets.

	\begin{table}[h]
		\centering
		\begin{threeparttable}[h]
			\footnotesize
			\caption{Ablation study on PGI.}
			\label{table:scale}
			\setlength\tabcolsep{5.5pt}
			\begin{tabular}{lcccccc}
				\toprule
				\textbf{Model} & \textbf{AP$^{val}_{50:95}$} &  & \textbf{AP$^{val}_{50}$} &  & \textbf{AP$^{val}_{75}$} &  \\
				\textbf{GELAN-S} & 46.7\% &  & 63.0\% &  & \textbf{50.7\%} &  \\
				\textbf{+ DS} & 46.5\% & -0.2 & 62.9\% & -0.1 & 50.5\% & -0.2 \\
				\textbf{+ PGI} & \textbf{46.8\%} & +0.1 & \textbf{63.4\%} & +0.4 & \textbf{50.7\%} & = \\
				\midrule
				\textbf{GELAN-M} & 51.1\% &  & 67.9\% &  & 55.7\% &  \\
				\textbf{+ DS} & 51.2\% & +0.1 & \textbf{68.2\%} & +0.3 & 55.7\% & = \\
				\textbf{+ PGI} & \textbf{51.4\%} & +0.3 & 68.1\% & +0.2 & \textbf{56.1\%} & +0.4 \\
				\midrule
				\textbf{GELAN-C} & 52.5\% &  & 69.5\% &  & 57.3\% &  \\
				\textbf{+ DS} & 52.5\% & = & 69.9\% & +0.4 & 57.1\% & -0.2 \\
				\textbf{+ PGI} & \textbf{53.0\%} & +0.5 & \textbf{70.3\%} & +0.8 & \textbf{57.8\%} & +0.5 \\
				\midrule
				\textbf{GELAN-E} & 55.0\% &  & 71.9\% &  & 60.0\% &  \\
				\textbf{+ DS} & 55.3\% & +0.3 & 72.3\% & +0.4 & 60.2\% & +0.2 \\
				\textbf{+ PGI} & \textbf{55.6\%} & +0.6 & \textbf{72.8\%} & +0.9 & \textbf{60.6\%} & +0.6 \\
				\bottomrule
			\end{tabular}
			\begin{tablenotes}[flushleft]
				\footnotesize
				\item[1] DS indicates deep supervision.
				\item[2] -\{S, M, C, E\} indicate small, medium, compact, and extended models.
			\end{tablenotes}
		\end{threeparttable}
	\end{table}

	\newpage
	
	Finally, we show in the table the results of gradually increasing components from baseline YOLOv7 to YOLOv9-E.  The GELAN and PGI we proposed have brought all-round improvement to the model.

	\begin{table}[h]
		\centering
		\begin{threeparttable}[h]
			\footnotesize
			\caption{Ablation study on GELAN and PGI.}
			\label{table:yolo}
			\setlength\tabcolsep{3.0pt}
			\begin{tabular}{lcccccc}
				\toprule
				\textbf{Model} & \textbf{\#Param.} & \textbf{FLOPs} & \textbf{AP$^{val}_{50:95}$} & \textbf{AP$^{val}_{S}$} & \textbf{AP$^{val}_{M}$} & \textbf{AP$^{val}_{L}$} \\
				\textbf{YOLOv7~\cite{wang2023yolov7}} & 36.9 & 104.7 & 51.2\% & 31.8\% & 55.5\% & 65.0\% \\
				\textbf{+ AF~\cite{wang2023yolov7}} & 43.6 & 130.5 & 53.0\% & 35.8\% & 58.7\% & 68.9\% \\
				\textbf{+ GELAN} & 41.2 & 126.4 & 53.2\% & 36.2\% & 58.5\% & 69.9\% \\
				\textbf{+ DHLC~\cite{liang2021cbnetv2}} & 57.3 & 189.0 & 55.0\% & 38.0\% & 60.6\% & 70.9\% \\
				\textbf{+ PGI} & 57.3 & 189.0 & 55.6\% & 40.2\% & 61.0\% & 71.4\% \\
				\bottomrule
			\end{tabular}
		\end{threeparttable}
	\end{table}

	\subsection{Visualization}
	
\begin{figure*}[t]
		\begin{center}
			\includegraphics[width=1.\linewidth]{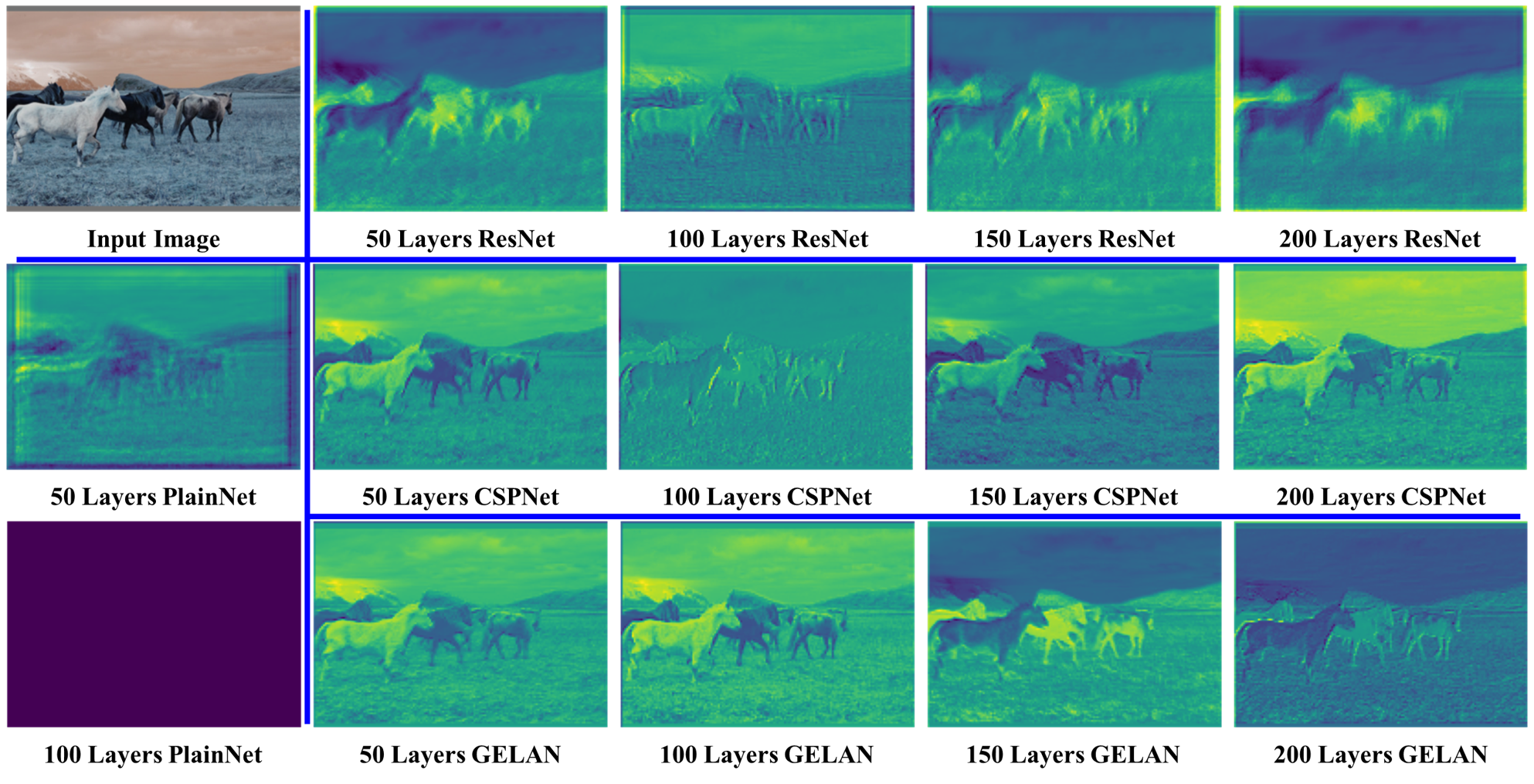}
		\end{center}
		\vspace{-14pt}
		\caption{Feature maps (visualization results) output by random initial weights of PlainNet, ResNet, CSPNet, and GELAN at different depths.  After 100 layers, ResNet begins to produce feedforward output that is enough to obfuscate object information.  Our proposed GELAN can still retain quite complete information up to the 150$^{th}$ layer, and is still sufficiently discriminative up to the 200$^{th}$ layer.}	
		\label{fig:deep}
		\vspace{-16pt}
	\end{figure*}

	This section will explore the information bottleneck issues and visualize them.  In addition, we will also visualize how the proposed PGI uses reliable gradients to find the correct correlations between data and targets.  In Figure~\ref{fig:deep} we show the visualization results of feature maps obtained by using random initial weights as feedforward under different architectures.  We can see that as the number of layers increases, the original information of all architectures gradually decreases.  For example, at the 50$^{th}$ layer of the PlainNet, it is difficult to see the location of objects, and all distinguishable features will be lost at the 100$^{th}$ layer.  As for ResNet, although the position of object can still be seen at the 50$^{th}$ layer, the boundary information has been lost.  When the depth reached to the 100$^{th}$ layer, the whole image becomes blurry.  Both CSPNet and the proposed GELAN perform very well, and they both can maintain features that support clear identification of objects until the 200$^{th}$ layer.  Among the comparisons, GELAN has more stable results and clearer boundary information.
	
	\begin{figure}[t]
		\begin{center}
			\includegraphics[width=1.\linewidth]{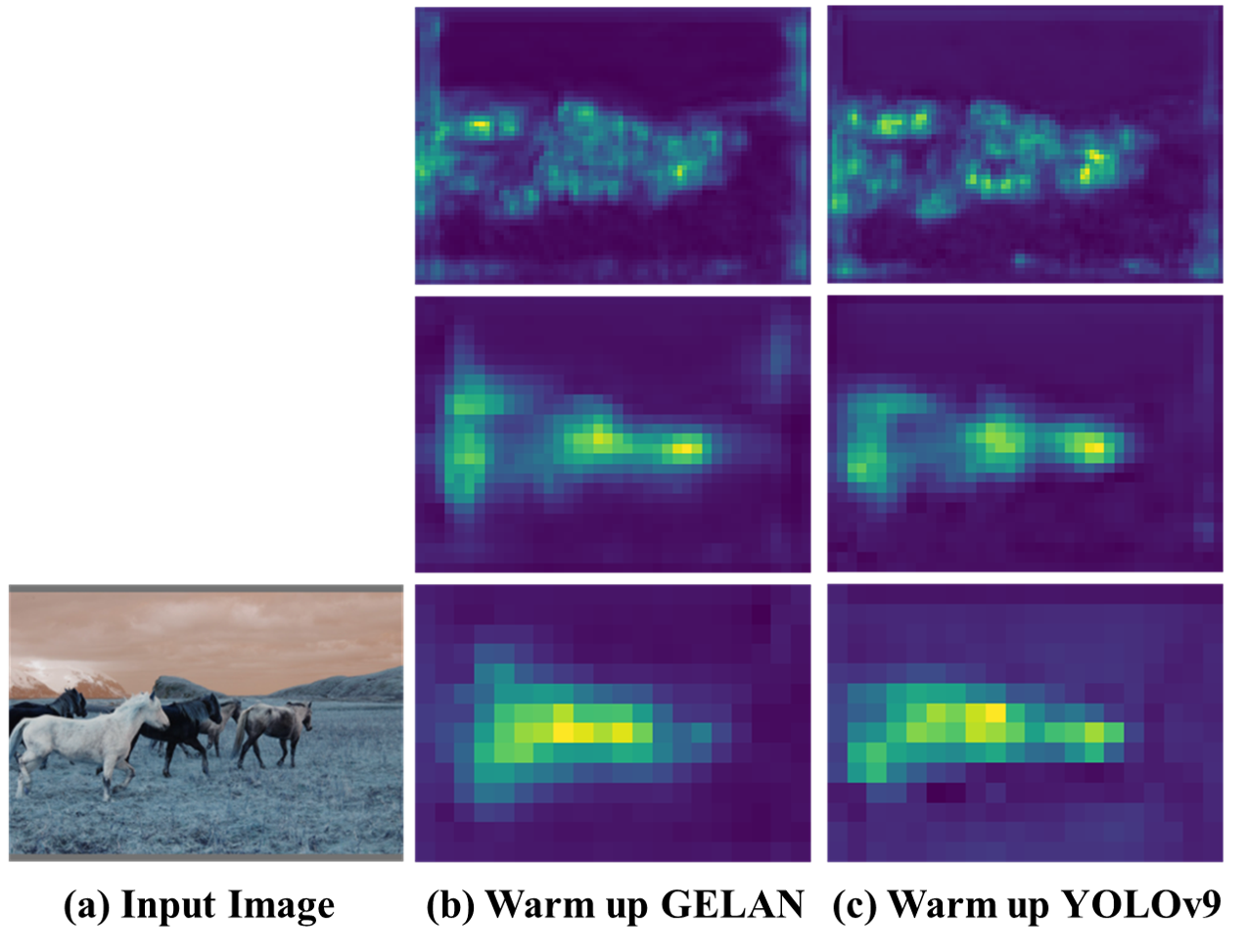}
		\end{center}
		\vspace{-12pt}
		\caption{PAN feature maps (visualization results) of GELAN and YOLOv9 (GELAN + PGI) after one epoch of bias warm-up.  GELAN originally had some divergence, but after adding PGI's reversible branch, it is more capable of focusing on the target object.}	
		\label{fig:warm}
		\vspace{-20pt}
	\end{figure}
	
	\newpage
	
	Figure~\ref{fig:warm} is used to show whether PGI can provide more reliable gradients during the training process, so that the parameters used for updating can effectively capture the relationship between the input data and the target.  Figure~\ref{fig:warm} shows the visualization results of the feature map of GELAN and YOLOv9 (GELAN + PGI) in PAN bias warm-up.  From the comparison of Figure~\ref{fig:warm}(b) and (c), we can clearly see that PGI accurately and concisely captures the area containing objects.  As for GELAN that does not use PGI, we found that it had divergence when detecting object boundaries, and it also produced unexpected responses in some background areas.  This experiment confirms that PGI can indeed provide better gradients to update parameters and enable the feedforward stage of the main branch to retain more important features.
	
	\section{Conclusions}
	
	In this paper, we propose to use PGI to solve the information bottleneck problem and the problem that the deep supervision mechanism is not suitable for lightweight neural networks.  We designed GELAN, a highly efficient and lightweight neural network.  In terms of object detection, GELAN has strong and stable performance at different computational blocks and depth settings.  It can indeed be widely expanded into a model suitable for various inference devices.  For the above two issues, the introduction of PGI allows both lightweight models and deep models to achieve significant improvements in accuracy.  The YOLOv9, designed by combining PGI and GELAN, has shown strong competitiveness.  Its excellent design allows the deep model to reduce the number of parameters by 49\% and the amount of calculations by 43\% compared with YOLOv8, but it still has a 0.6\% AP improvement on MS COCO dataset.
	
	\section{Acknowledgements}
	
	The authors wish to thank National Center for High-performance Computing (NCHC) for providing computational and storage resources.
	
	\clearpage
	\clearpage
	\clearpage
	
	
	\clearpage
	\clearpage
	\clearpage

	\appendix
	
	\setcounter{page}{1}
	\setcounter{table}{0}
	\setcounter{figure}{0}
	
	\twocolumn[
	\centering
	\Large
	\textbf{Appendix} \\
	\vspace{1.0em}
	]
	
	\appendix
	
	\section{Implementation Details}
	
	\begin{table}[h]
		\centering
		\begin{threeparttable}[h]
			\caption{Hyper parameter settings of YOLOv9.}
			\label{table:hyp}
			\begin{tabular}{lc}
				\toprule
				\textbf{hyper parameter} & \textbf{value} \\	
				\midrule
				epochs & 500 \\
				optimizer & SGD \\
				initial learning rate & 0.01 \\
				finish learning rate & 0.0001 \\
				learning rate decay & linear \\
				momentum & 0.937 \\
				weight decay & 0.0005 \\
				warm-up epochs & 3 \\
				warm-up momentum & 0.8 \\
				warm-up bias learning rate & 0.1 \\
				box loss gain & 7.5 \\
				class loss gain & 0.5 \\
				DFL loss gain & 1.5 \\
				HSV saturation augmentation & 0.7 \\
				HSV value augmentation & 0.4 \\
				translation augmentation & 0.1 \\
				scale augmentation & 0.9 \\
				mosaic augmentation & 1.0 \\
				MixUp augmentation & 0.15 \\
				copy \& paste augmentation & 0.3 \\
				close mosaic epochs & 15 \\
				\bottomrule
			\end{tabular}
		\end{threeparttable}
	\end{table}
	
	The training parameters of YOLOv9 are shown in Table~\ref{table:hyp}.  We fully follow the settings of YOLOv7 AF~\cite{wang2023yolov7}, which is to use SGD optimizer to train 500 epochs.  We first warm-up for 3 epochs and only update the bias during the warm-up stage.  Next we step down from the initial learning rate 0.01 to 0.0001 in linear decay manner, and the data augmentation settings are listed in the bottom part of Table~\ref{table:hyp}.  We shut down mosaic data augmentation operations on the last 15 epochs.
	
	\newpage
	
	\begin{table}[h]
		\centering
		\begin{threeparttable}[h]
			\small
			\caption{Network configurations of YOLOv9.}
			\label{table:cfg}
			\setlength\tabcolsep{1.0pt}
			\begin{tabular}{lcccccc}
				\toprule
				\textbf{Index} & \textbf{Module} & \textbf{Route} & \textbf{Filters} & \textbf{Depth} & \textbf{Size} & \textbf{Stride} \\	
				0 & Conv & -- & 64 & -- & 3 & 2 \\	
				1 & Conv & 0 & 128 & -- & 3 & 2 \\	
				2 & CSP-ELAN & 1 & 256, 128, 64 & 2, 1 & -- & 1 \\	
				3 & DOWN & 2 & 256 & -- & 3 & 2 \\	
				4 & CSP-ELAN & 3 & 512, 256, 128 & 2, 1 & -- & 1 \\	
				5 & DOWN & 4 & 512 & -- & 3 & 2 \\
				6 & CSP-ELAN & 5 & 512, 512, 256 & 2, 1 & -- & 1 \\	
				7 & DOWN & 6 & 512 & -- & 3 & 2 \\
				8 & CSP-ELAN & 7 & 512, 512, 256 & 2, 1 & -- & 1 \\		
				9 & SPP-ELAN & 8 & 512, 256, 256 & 3, 1 & -- & 1 \\	
				10 & Up & 9 & 512 & -- & -- & 2 \\	
				11 & Concat & 10, 6 & 1024 & -- & -- & 1 \\	
				12 & CSP-ELAN & 11 & 512, 512, 256 & 2, 1 & -- & 1 \\	
				13 & Up & 12 & 512 & -- & -- & 2 \\	
				14 & Concat & 13, 4 & 1024 & -- & -- & 1 \\
				15 & CSP-ELAN & 14 & 256, 256, 128 & 2, 1 & -- & 1 \\	
				16 & DOWN & 15 & 256 & -- & 3 & 2 \\	
				17 & Concat & 16, 12 & 768 & -- & -- & 1 \\	
				18 & CSP-ELAN & 17 & 512, 512, 256 & 2, 1 & -- & 1 \\	
				19 & DOWN & 18 & 512 & -- & 3 & 2 \\	
				20 & Concat & 19, 9 & 1024 & -- & -- & 1 \\
				21 & CSP-ELAN & 20 & 512, 512, 256 & 2, 1 & -- & 1 \\	
				22 & Predict & 15, 18, 21 & -- & -- & -- & -- \\	
				\bottomrule
			\end{tabular}
		\end{threeparttable}
	\end{table}
	
	The network topology of YOLOv9 completely follows YOLOv7 AF~\cite{wang2023yolov7}, that is, we replace ELAN with the proposed CSP-ELAN block.  As listed in Table~\ref{table:cfg}, the depth parameters of CSP-ELAN are represented as ELAN depth and CSP depth, respectively.  As for the parameters of CSP-ELAN filters, they are represented as ELAN output filter, CSP output filter, and CSP inside filter.  In the down-sampling module part, we simplify CSP-DOWN module to DOWN module.  DOWN module is composed of a pooling layer with size 2 and stride 1, and a Conv layer with size 3 and stride 2.  Finally, we optimized the prediction layer and replaced top, left, bottom, and right in the regression branch with decoupled branch.
	
	\begin{table*}[h]
		\centering
		\begin{threeparttable}[t]
			\footnotesize
			\caption{Comparison of state-of-the-art object detectors with different training settings.}
			\label{table:more}
			\setlength\tabcolsep{4.0pt}
			\begin{tabular}{l|lcccccccc}
				\toprule
				& \textbf{Model} & \textbf{\#Param. (M)} & \textbf{FLOPs (G)} & \textbf{AP$_{50:95}$ (\%)} & \textbf{AP$_{50}$ (\%)} & \textbf{AP$_{75}$ (\%)} & \textbf{AP$_{S}$ (\%)} & \textbf{AP$_{M}$ (\%)} & \textbf{AP$_{L}$ (\%)} \\
				\midrule
				\multirow{8}{*}{\rotatebox{90}{\textbf{Train-from-scratch}}} & \textbf{Dy-YOLOv7~\cite{lin2023dynamicdet}} & -- & 181.7 & 53.9 & 72.2 & 58.7 & 35.3 & 57.6 & 66.4 \\
				& \textbf{Dy-YOLOv7-X~\cite{lin2023dynamicdet}} & -- & 307.9 & 55.0 & 73.2 & 60.0 & 36.6 & 58.7 & 68.5 \\
				\cline{2-10}
				& \textbf{YOLOv9-S (Ours)} & 7.1 & 26.4 & 46.8 & 63.4 & 50.7 & 26.6 & 56.0 & 64.5 \\
				& \textbf{YOLOv9-M (Ours)} & 20.0 & 76.3 & 51.4 & 68.1 & 56.1 & 33.6 & 57.0 & 68.0 \\
				& \textbf{YOLOv9-C (Ours)} & 25.3 & 102.1 & 53.0 & 70.2 & 57.8 & 36.2 & 58.5 & 69.3 \\
				& \textbf{YOLOv9-E (Ours)} & 34.7 & 147.1 & 54.5 & 71.7 & 59.2 & 38.1 & 59.9 & 70.3 \\
				& \textbf{YOLOv9-E (Ours)} & 44.0 & 183.9 & 55.1 & 72.3 & 60.7 & 38.7 & 60.6 & 71.4 \\
				& \textbf{YOLOv9-E (Ours)} & 57.3 & 189.0 & 55.6 & 72.8 & 60.6 & 40.2 & 61.0 & 71.4 \\
				\midrule
				\multirow{19}{*}{\rotatebox{90}{\textbf{ImageNet Pretrained}}} & \textbf{RTMDet-T~\cite{lyu2022rtmdet}} & 4.8 & 12.6 & 41.1 & 57.9 & -- & -- & -- & -- \\
				& \textbf{RTMDet-S~\cite{lyu2022rtmdet}} & 9.0 & 25.6 & 44.6 & 61.9 & -- & -- & -- & -- \\
				& \textbf{RTMDet-M~\cite{lyu2022rtmdet}} & 24.7 & 78.6 & 49.4 & 66.8 & -- & -- & -- & -- \\
				& \textbf{RTMDet-L~\cite{lyu2022rtmdet}} & 52.3 & 160.4 & 51.5 & 68.8 & -- & -- & -- & -- \\
				& \textbf{RTMDet-X~\cite{lyu2022rtmdet}} & 94.9 & 283.4 & 52.8 & 70.4 & -- & -- & -- & -- \\
				\cline{2-10}
				& \textbf{PPYOLOE-S~\cite{xu2022pp}} & 7.9 & 14.4 & 43.0 & 60.5 & 46.6 & 23.2 & 46.4 & 56.9 \\
				& \textbf{PPYOLOE-M~\cite{xu2022pp}} & 23.4 & 49.9 & 49.0 & 66.5 & 53.0 & 28.6 & 52.9 & 63.8 \\
				& \textbf{PPYOLOE-L~\cite{xu2022pp}} & 52.2 & 110.1 & 51.4 & 68.9 & 55.6 & 31.4 & 55.3 & 66.1 \\
				& \textbf{PPYOLOE-X~\cite{xu2022pp}} & 98.4 & 206.6 & 52.3 & 69.5 & 56.8 & 35.1 & 57.0 & 68.6 \\
				\cline{2-10}
				& \textbf{RT DETR-L~\cite{lv2023detrs}} & 32 & 110 & 53.0 & 71.6 & 57.3 & 34.6 & 57.3 & 71.2 \\
				& \textbf{RT DETR-X~\cite{lv2023detrs}} & 67 & 234 & 54.8 & 73.1 & 59.4 & 35.7 & 59.6 & 72.9 \\
				& \textbf{RT DETR-R18~\cite{lv2023detrs}} & 20 & 60 & 46.5 & 63.8 & -- & -- & -- & -- \\
				& \textbf{RT DETR-R34~\cite{lv2023detrs}} & 31 & 92 & 48.9 & 66.8 & -- & -- & -- & -- \\
				& \textbf{RT DETR-R50M~\cite{lv2023detrs}} & 36 & 100 & 51.3 & 69.6 & -- & -- & -- & -- \\
				& \textbf{RT DETR-R50~\cite{lv2023detrs}} & 42 & 136 & 53.1 & 71.3 & 57.7 & 34.8 & 58.0 & 70.0 \\
				& \textbf{RT DETR-R101~\cite{lv2023detrs}} & 76 & 259 & 54.3 & 72.7 & 58.6 & 36.0 & 58.8 & 72.1 \\
				\cline{2-10}
				& \textbf{Gold YOLO-S~\cite{wang2023gold}} & 21.5 & 46.0 & 45.5 & 62.2 & -- & -- & -- & -- \\
				& \textbf{Gold YOLO-M~\cite{wang2023gold}} & 41.3 & 57.5 & 50.2 & 67.5 & -- & -- & -- & -- \\
				& \textbf{Gold YOLO-L~\cite{wang2023gold}} & 75.1 & 151.7 & 52.3 & 69.6 & -- & -- & -- & -- \\
				\midrule
				\multirow{12}{*}{\rotatebox{90}{\textbf{Knowledge Distillation}}} & \textbf{YOLOv6-N v3.0~\cite{li2023yolov6}} & 4.7 & 11.4 & 37.5 & 53.1 & -- & -- & -- & -- \\
				& \textbf{YOLOv6-S v3.0~\cite{li2023yolov6}} & 18.5 & 45.3 & 45.0 & 61.8 & -- & -- & -- & -- \\
				& \textbf{YOLOv6-M v3.0~\cite{li2023yolov6}} & 34.9 & 85.8 & 50.0 & 66.9 & -- & -- & -- & -- \\
				& \textbf{YOLOv6-L v3.0~\cite{li2023yolov6}} & 59.6 & 150.7 & 52.8 & 70.3 & -- & -- & -- & -- \\
				\cline{2-10}
				& \textbf{DAMO YOLO-T~\cite{xu2022damo}} & 8.5 & 18.1 & 43.6 & 59.4 & 46.6 & 23.3 & 47.4 & 61.0 \\
				& \textbf{DAMO YOLO-S~\cite{xu2022damo}} & 16.3 & 37.8 & 47.7 & 63.5 & 51.1 & 26.9 & 51.7 & 64.9 \\
				& \textbf{DAMO YOLO-M~\cite{xu2022damo}} & 28.2 & 61.8 & 50.4 & 67.2 & 55.1 & 31.6 & 55.3 & 67.1 \\
				& \textbf{DAMO YOLO-L~\cite{xu2022damo}} & 42.1 & 97.3 & 51.9 & 68.5 & 56.7 & 33.3 & 57.0 & 67.6 \\
				\cline{2-10}
				& \textbf{Gold YOLO-N~\cite{wang2023gold}} & 5.6 & 12.1 & 39.9 & 55.9 & -- & -- & -- & -- \\
				& \textbf{Gold YOLO-S~\cite{wang2023gold}} & 21.5 & 46.0 & 46.1 & 63.3 & -- & -- & -- & -- \\
				& \textbf{Gold YOLO-M~\cite{wang2023gold}} & 41.3 & 57.5 & 50.9 & 68.2 & -- & -- & -- & -- \\
				& \textbf{Gold YOLO-L~\cite{wang2023gold}} & 75.1 & 151.7 & 53.2 & 70.5 & -- & -- & -- & -- \\
				\midrule
				\multirow{10}{*}{\rotatebox{90}{\textbf{Complex Setting}}} & \textbf{Gold YOLO-S~\cite{wang2023gold}} & 21.5 & 46.0 & 46.4 & 63.4 & -- & -- & -- & -- \\
				& \textbf{Gold YOLO-M~\cite{wang2023gold}} & 41.3 & 57.5 & 51.1 & 68.5 & -- & -- & -- & -- \\
				& \textbf{Gold YOLO-L~\cite{wang2023gold}} & 75.1 & 151.7 & 53.3 & 70.9 & -- & -- & -- & -- \\
				\cline{2-10}				
				& \textbf{YOLOR-CSP~\cite{wang2021you}} & 52.9 & 120.4 & 52.8 & 71.2 & 57.6 & -- & -- & -- \\
				& \textbf{YOLOR-CSP-X~\cite{wang2021you}} & 96.9 & 226.8 & 54.8 & 73.1 & 59.7 &--  & -- & -- \\
				\cline{2-10}
				& \textbf{PPYOLOE+-S~\cite{xu2022pp}} & 7.9 & 14.4 & 43.7 & 60.6 & 47.9 & 23.2 & 46.4 & 56.9 \\
				& \textbf{PPYOLOE+-M~\cite{xu2022pp}} & 23.4 & 49.9 & 49.8 & 67.1 & 54.5 & 31.8 & 53.9 & 66.2 \\
				& \textbf{PPYOLOE+-L~\cite{xu2022pp}} & 52.2 & 110.1 & 52.9 & 70.1 & 57.9 & 35.2 & 57.5 & 69.1 \\
				& \textbf{PPYOLOE+-X~\cite{xu2022pp}} & 98.4 & 206.6 & 54.7 & 72.0 & 59.9 & 37.9 & 59.3 & 70.4 \\
				\bottomrule
			\end{tabular}
		\end{threeparttable}
		\vspace{-4pt}
	\end{table*}
	
	\newpage
	
	\section{More Comparison}
	
	\vspace{-8pt}
	
	We compare YOLOv9 to state-of-the-art real-time object detectors trained with different methods.  It mainly includes four different training methods: (1) train-from-scratch: we have completed most of the comparisons in the text.  Here are only list of additional data of DynamicDet~\cite{lin2023dynamicdet} for comparisons; (2) Pretrained by ImageNet: this includes two methods of using ImageNet for supervised pretrain and self-supervised pretrain; (3) knowledge distillation: a method to perform additional self-distillation after training is completed; and (4) a more complex training process: a combination of steps including pretrained by ImageNet, knowledge distillation, DAMO-YOLO and even additional pretrained large object detection dataset.  We show the results in Table~\ref{table:more}.  From this table, we can see that our proposed YOLOv9 performed better than all other methods.  Compared with PPYOLOE+-X trained using ImageNet and Objects365, our method still reduces the number of parameters by 55\% and the amount of computation by 11\%, and improving 0.4\% AP.
	
	\begin{table*}[t]
		\centering
		\begin{threeparttable}[t]
			\footnotesize
			\caption{Comparison of state-of-the-art object detectors with different training settings (sorted by number of parameters).}
			\label{table:param}
			\setlength\tabcolsep{4.0pt}
			\begin{tabular}{l|lcccccccc}
				\toprule
				& \textbf{Model} & \textbf{\#Param. (M)} & \textbf{FLOPs (G)} & \textbf{AP$^{val}_{50:95}$ (\%)} & \textbf{AP$^{val}_{50}$ (\%)} & \textbf{AP$^{val}_{75}$ (\%)} & \textbf{AP$^{val}_{S}$ (\%)} & \textbf{AP$^{val}_{M}$ (\%)} & \textbf{AP$^{val}_{L}$ (\%)} \\
				\midrule
				& \textbf{YOLOv6-N v3.0~\cite{li2023yolov6} (D)} & 4.7 & 11.4 & \textbf{37.5} & 53.1 & -- & -- & -- & -- \\
				& \textbf{RTMDet-T~\cite{lyu2022rtmdet} (I)} & 4.8 & 12.6 & \textbf{41.1} & 57.9 & -- & -- & -- & -- \\
				& \textbf{Gold YOLO-N~\cite{wang2023gold} (D)} & 5.6 & 12.1 & 39.9 & 55.9 & -- & -- & -- & -- \\
				& \textbf{YOLOv9-S (S)} & 7.1 & 26.4 &\textbf{ 46.8} & 63.4 & 50.7 & 26.6 & 56.0 & 64.5 \\
				& \textbf{PPYOLOE+-S~\cite{xu2022pp} (C)} & 7.9 & 14.4 & 43.7 & 60.6 & 47.9 & 23.2 & 46.4 & 56.9 \\
				& \textbf{PPYOLOE-S~\cite{xu2022pp} (I)} & 7.9 & 14.4 & 43.0 & 60.5 & 46.6 & 23.2 & 46.4 & 56.9 \\
				& \textbf{DAMO YOLO-T~\cite{xu2022damo} (D)} & 8.5 & 18.1 & 43.6 & 59.4 & 46.6 & 23.3 & 47.4 & 61.0 \\
				& \textbf{RTMDet-S~\cite{lyu2022rtmdet} (I)} & 9.0 & 25.6 & 44.6 & 61.9 & -- & -- & -- & -- \\
				& \textbf{DAMO YOLO-S~\cite{xu2022damo} (D)} & 16.3 & 37.8 & \textbf{47.7} & 63.5 & 51.1 & 26.9 & 51.7 & 64.9 \\
				& \textbf{YOLOv6-S v3.0~\cite{li2023yolov6} (D)} & 18.5 & 45.3 & 45.0 & 61.8 & -- & -- & -- & -- \\
				& \textbf{RT DETR-R18~\cite{lv2023detrs} (I)} & 20 & 60 & 46.5 & 63.8 & -- & -- & -- & -- \\
				& \textbf{YOLOv9-M (S)} & 20.0 & 76.3 & \textbf{51.4} & 68.1 & 56.1 & 33.6 & 57.0 & 68.0 \\
				& \textbf{Gold YOLO-S~\cite{wang2023gold} (C)} & 21.5 & 46.0 & 46.4 & 63.4 & -- & -- & -- & -- \\
				& \textbf{Gold YOLO-S~\cite{wang2023gold} (D)} & 21.5 & 46.0 & 46.1 & 63.3 & -- & -- & -- & -- \\
				& \textbf{Gold YOLO-S~\cite{wang2023gold} (I)} & 21.5 & 46.0 & 45.5 & 62.2 & -- & -- & -- & -- \\
				& \textbf{PPYOLOE+-M~\cite{xu2022pp} (C)} & 23.4 & 49.9 & 49.8 & 67.1 & 54.5 & 31.8 & 53.9 & 66.2 \\
				& \textbf{PPYOLOE-M~\cite{xu2022pp} (I)} & 23.4 & 49.9 & 49.0 & 66.5 & 53.0 & 28.6 & 52.9 & 63.8 \\
				& \textbf{RTMDet-M~\cite{lyu2022rtmdet} (I)} & 24.7 & 78.6 & 49.4 & 66.8 & -- & -- & -- & -- \\
				& \textbf{YOLOv9-C (S)} & 25.3 & 102.1 & \textbf{53.0} & 70.2 & 57.8 & 36.2 & 58.5 & 69.3 \\
				& \textbf{DAMO YOLO-M~\cite{xu2022damo} (D)} & 28.2 & 61.8 & 50.4 & 67.2 & 55.1 & 31.6 & 55.3 & 67.1 \\
				& \textbf{RT DETR-R34~\cite{lv2023detrs} (I)} & 31 & 92 & 48.9 & 66.8 & -- & -- & -- & -- \\
				& \textbf{RT DETR-L~\cite{lv2023detrs} (I)} & 32 & 110 & 53.0 & 71.6 & 57.3 & 34.6 & 57.3 & 71.2 \\
				& \textbf{YOLOv9-E (S)} & 34.7 & 147.1 & \textbf{54.5} & 71.7 & 59.2 & 38.1 & 59.9 & 70.3 \\
				& \textbf{YOLOv6-M v3.0~\cite{li2023yolov6} (D)} & 34.9 & 85.8 & 50.0 & 66.9 & -- & -- & -- & -- \\
				& \textbf{RT DETR-R50M~\cite{lv2023detrs} (I)} & 36 & 100 & 51.3 & 69.6 & -- & -- & -- & -- \\
				& \textbf{Gold YOLO-M~\cite{wang2023gold} (C)} & 41.3 & 57.5 & 51.1 & 68.5 & -- & -- & -- & -- \\
				& \textbf{Gold YOLO-M~\cite{wang2023gold} (D)} & 41.3 & 57.5 & 50.9 & 68.2 & -- & -- & -- & -- \\
				& \textbf{Gold YOLO-M~\cite{wang2023gold} (I)} & 41.3 & 57.5 & 50.2 & 67.5 & -- & -- & -- & -- \\
				& \textbf{RT DETR-R50~\cite{lv2023detrs} (I)} & 42 & 136 & 53.1 & 71.3 & 57.7 & 34.8 & 58.0 & 70.0 \\
				& \textbf{DAMO YOLO-L~\cite{xu2022damo} (D)} & 42.1 & 97.3 & 51.9 & 68.5 & 56.7 & 33.3 & 57.0 & 67.6 \\
				& \textbf{YOLOv9-E (S)} & 44.0 & 183.9 & \textbf{55.1} & 72.3 & 60.7 & 38.7 & 60.6 & 71.4 \\
				& \textbf{PPYOLOE+-L~\cite{xu2022pp} (C)} & 52.2 & 110.1 & 52.9 & 70.1 & 57.9 & 35.2 & 57.5 & 69.1 \\
				& \textbf{PPYOLOE-L~\cite{xu2022pp} (I)} & 52.2 & 110.1 & 51.4 & 68.9 & 55.6 & 31.4 & 55.3 & 66.1 \\
				& \textbf{RTMDet-L~\cite{lyu2022rtmdet} (I)} & 52.3 & 160.4 & 51.5 & 68.8 & -- & -- & -- & -- \\
				& \textbf{YOLOR-CSP~\cite{wang2021you} (C)} & 52.9 & 120.4 & 52.8 & 71.2 & 57.6 & -- & -- & -- \\
				& \textbf{YOLOv9-E (S)} & 57.3 & 189.0 & \textbf{55.6} & 72.8 & 60.6 & 40.2 & 61.0 & 71.4 \\
				& \textbf{YOLOv6-L v3.0~\cite{li2023yolov6} (D)} & 59.6 & 150.7 & 52.8 & 70.3 & -- & -- & -- & -- \\
				& \textbf{RT DETR-X~\cite{lv2023detrs} (I)} & 67 & 234 & 54.8 & 73.1 & 59.4 & 35.7 & 59.6 & 72.9 \\
				& \textbf{Gold YOLO-L~\cite{wang2023gold} (C)} & 75.1 & 151.7 & 53.3 & 70.9 & -- & -- & -- & -- \\
				& \textbf{Gold YOLO-L~\cite{wang2023gold} (D)} & 75.1 & 151.7 & 53.2 & 70.5 & -- & -- & -- & -- \\	
				& \textbf{Gold YOLO-L~\cite{wang2023gold} (I)} & 75.1 & 151.7 & 52.3 & 69.6 & -- & -- & -- & -- \\
				& \textbf{RT DETR-R101~\cite{lv2023detrs} (I)} & 76 & 259 & 54.3 & 72.7 & 58.6 & 36.0 & 58.8 & 72.1 \\
				& \textbf{RTMDet-X~\cite{lyu2022rtmdet} (I)} & 94.9 & 283.4 & 52.8 & 70.4 & -- & -- & -- & -- \\
				& \textbf{YOLOR-CSP-X~\cite{wang2021you} (C)} & 96.9 & 226.8 & 54.8 & 73.1 & 59.7 &--  & -- & -- \\
				& \textbf{PPYOLOE+-X~\cite{xu2022pp} (C)} & 98.4 & 206.6 & 54.7 & 72.0 & 59.9 & 37.9 & 59.3 & 70.4 \\	
				& \textbf{PPYOLOE-X~\cite{xu2022pp} (I)} & 98.4 & 206.6 & 52.3 & 69.5 & 56.8 & 35.1 & 57.0 & 68.6 \\
				\bottomrule
			\end{tabular}
			\begin{tablenotes}[flushleft]
			\footnotesize
			\item[1] (S), (I), (D), (C) indicate train-from-scratch, ImageNet pretrained, knowledge distillation, and complex setting, respectively.
			\end{tablenotes}
		\end{threeparttable}
	\end{table*}
	
	\newpage
	
	Table~\ref{table:param} shows the performance of all models sorted by parameter size. Our proposed YOLOv9 is Pareto optimal in all models of different sizes. Among them, we found no other method for Pareto optimal in models with more than 20M parameters. The above experimental data shows that our YOLOv9 has excellent parameter usage efficiency.
	
	\newpage
	
	\begin{table*}[t]
		\centering
		\begin{threeparttable}[t]
			\footnotesize
			\caption{Comparison of state-of-the-art object detectors with different training settings (sorted by amount of computation).}
			\label{table:flops}
			\setlength\tabcolsep{4.0pt}
			\begin{tabular}{l|lcccccccc}
				\toprule
				& \textbf{Model} & \textbf{\#Param. (M)} & \textbf{FLOPs (G)} & \textbf{AP$^{val}_{50:95}$ (\%)} & \textbf{AP$^{val}_{50}$ (\%)} & \textbf{AP$^{val}_{75}$ (\%)} & \textbf{AP$^{val}_{S}$ (\%)} & \textbf{AP$^{val}_{M}$ (\%)} & \textbf{AP$^{val}_{L}$ (\%)} \\
				\midrule
				& \textbf{YOLOv6-N v3.0~\cite{li2023yolov6} (D)} & 4.7 & 11.4 & \textbf{37.5} & 53.1 & -- & -- & -- & -- \\
				& \textbf{Gold YOLO-N~\cite{wang2023gold} (D)} & 5.6 & 12.1 & \textbf{39.9} & 55.9 & -- & -- & -- & -- \\
				& \textbf{RTMDet-T~\cite{lyu2022rtmdet} (I)} & 4.8 & 12.6 & \textbf{41.1} & 57.9 & -- & -- & -- & -- \\
				& \textbf{PPYOLOE+-S~\cite{xu2022pp} (C)} & 7.9 & 14.4 & \textbf{43.7} & 60.6 & 47.9 & 23.2 & 46.4 & 56.9 \\
				& \textbf{PPYOLOE-S~\cite{xu2022pp} (I)} & 7.9 & 14.4 & 43.0 & 60.5 & 46.6 & 23.2 & 46.4 & 56.9 \\
				& \textbf{DAMO YOLO-T~\cite{xu2022damo} (D)} & 8.5 & 18.1 & 43.6 & 59.4 & 46.6 & 23.3 & 47.4 & 61.0 \\
				& \textbf{RTMDet-S~\cite{lyu2022rtmdet} (I)} & 9.0 & 25.6 & \textbf{44.6} & 61.9 & -- & -- & -- & -- \\
				& \textbf{YOLOv9-S (S)} & 7.1 & 26.4 & \textbf{46.8} & 63.4 & 50.7 & 26.6 & 56.0 & 64.5 \\
				& \textbf{DAMO YOLO-S~\cite{xu2022damo} (D)} & 16.3 & 37.8 & \textbf{47.7} & 63.5 & 51.1 & 26.9 & 51.7 & 64.9 \\
				& \textbf{YOLOv6-S v3.0~\cite{li2023yolov6} (D)} & 18.5 & 45.3 & 45.0 & 61.8 & -- & -- & -- & -- \\
				& \textbf{Gold YOLO-S~\cite{wang2023gold} (C)} & 21.5 & 46.0 & 46.4 & 63.4 & -- & -- & -- & -- \\
				& \textbf{Gold YOLO-S~\cite{wang2023gold} (D)} & 21.5 & 46.0 & 46.1 & 63.3 & -- & -- & -- & -- \\
				& \textbf{Gold YOLO-S~\cite{wang2023gold} (I)} & 21.5 & 46.0 & 45.5 & 62.2 & -- & -- & -- & -- \\
				& \textbf{PPYOLOE+-M~\cite{xu2022pp} (C)} & 23.4 & 49.9 & \textbf{49.8} & 67.1 & 54.5 & 31.8 & 53.9 & 66.2 \\
				& \textbf{PPYOLOE-M~\cite{xu2022pp} (I)} & 23.4 & 49.9 & 49.0 & 66.5 & 53.0 & 28.6 & 52.9 & 63.8 \\
				& \textbf{Gold YOLO-M~\cite{wang2023gold} (C)} & 41.3 & 57.5 & \textbf{51.1} & 68.5 & -- & -- & -- & -- \\
				& \textbf{Gold YOLO-M~\cite{wang2023gold} (D)} & 41.3 & 57.5 & 50.9 & 68.2 & -- & -- & -- & -- \\
				& \textbf{Gold YOLO-M~\cite{wang2023gold} (I)} & 41.3 & 57.5 & 50.2 & 67.5 & -- & -- & -- & -- \\
				& \textbf{RT DETR-R18~\cite{lv2023detrs} (I)} & 20 & 60 & 46.5 & 63.8 & -- & -- & -- & -- \\
				& \textbf{DAMO YOLO-M~\cite{xu2022damo} (D)} & 28.2 & 61.8 & 50.4 & 67.2 & 55.1 & 31.6 & 55.3 & 67.1 \\
				& \textbf{YOLOv9-M (S)} & 20.0 & 76.3 & \textbf{51.4} & 68.1 & 56.1 & 33.6 & 57.0 & 68.0 \\
				& \textbf{RTMDet-M~\cite{lyu2022rtmdet} (I)} & 24.7 & 78.6 & 49.4 & 66.8 & -- & -- & -- & -- \\
				& \textbf{YOLOv6-M v3.0~\cite{li2023yolov6} (D)} & 34.9 & 85.8 & 50.0 & 66.9 & -- & -- & -- & -- \\
				& \textbf{RT DETR-R34~\cite{lv2023detrs} (I)} & 31 & 92 & 48.9 & 66.8 & -- & -- & -- & -- \\
				& \textbf{DAMO YOLO-L~\cite{xu2022damo} (D)} & 42.1 & 97.3 & \textbf{51.9} & 68.5 & 56.7 & 33.3 & 57.0 & 67.6 \\
				& \textbf{RT DETR-R50M~\cite{lv2023detrs} (I)} & 36 & 100 & 51.3 & 69.6 & -- & -- & -- & -- \\
				& \textbf{YOLOv9-C (S)} & 25.3 & 102.1 & \textbf{53.0} & 70.2 & 57.8 & 36.2 & 58.5 & 69.3 \\
				& \textbf{RT DETR-L~\cite{lv2023detrs} (I)} & 32 & 110 & 53.0 & 71.6 & 57.3 & 34.6 & 57.3 & 71.2 \\
				& \textbf{PPYOLOE+-L~\cite{xu2022pp} (C)} & 52.2 & 110.1 & 52.9 & 70.1 & 57.9 & 35.2 & 57.5 & 69.1 \\
				& \textbf{PPYOLOE-L~\cite{xu2022pp} (I)} & 52.2 & 110.1 & 51.4 & 68.9 & 55.6 & 31.4 & 55.3 & 66.1 \\
				& \textbf{YOLOR-CSP~\cite{wang2021you} (C)} & 52.9 & 120.4 & 52.8 & 71.2 & 57.6 & -- & -- & -- \\
				& \textbf{RT DETR-R50~\cite{lv2023detrs} (I)} & 42 & 136 & \textbf{53.1} & 71.3 & 57.7 & 34.8 & 58.0 & 70.0 \\
				& \textbf{YOLOv9-E (S)} & 34.7 & 147.1 & \textbf{54.5} & 71.7 & 59.2 & 38.1 & 59.9 & 70.3 \\
				& \textbf{YOLOv6-L v3.0~\cite{li2023yolov6} (D)} & 59.6 & 150.7 & 52.8 & 70.3 & -- & -- & -- & -- \\
				& \textbf{Gold YOLO-L~\cite{wang2023gold} (C)} & 75.1 & 151.7 & 53.3 & 70.9 & -- & -- & -- & -- \\
				& \textbf{Gold YOLO-L~\cite{wang2023gold} (D)} & 75.1 & 151.7 & 53.2 & 70.5 & -- & -- & -- & -- \\	
				& \textbf{Gold YOLO-L~\cite{wang2023gold} (I)} & 75.1 & 151.7 & 52.3 & 69.6 & -- & -- & -- & -- \\
				& \textbf{RTMDet-L~\cite{lyu2022rtmdet} (I)} & 52.3 & 160.4 & 51.5 & 68.8 & -- & -- & -- & -- \\
				& \textbf{Dy-YOLOv7~\cite{lin2023dynamicdet} (S)} & -- & 181.7 & 53.9 & 72.2 & 58.7 & 35.3 & 57.6 & 66.4 \\
				& \textbf{YOLOv9-E (S)} & 44.0 & 183.9 & \textbf{55.1} & 72.3 & 60.7 & 38.7 & 60.6 & 71.4 \\
				& \textbf{YOLOv9-E (S)} & 57.3 & 189.0 & \textbf{55.6} & 72.8 & 60.6 & 40.2 & 61.0 & 71.4 \\
				& \textbf{PPYOLOE+-X~\cite{xu2022pp} (C)} & 98.4 & 206.6 & 54.7 & 72.0 & 59.9 & 37.9 & 59.3 & 70.4 \\
				& \textbf{PPYOLOE-X~\cite{xu2022pp} (I)} & 98.4 & 206.6 & 52.3 & 69.5 & 56.8 & 35.1 & 57.0 & 68.6 \\
				& \textbf{YOLOR-CSP-X~\cite{wang2021you} (C)} & 96.9 & 226.8 & 54.8 & 73.1 & 59.7 &--  & -- & -- \\
				& \textbf{RT DETR-X~\cite{lv2023detrs} (I)} & 67 & 234 & 54.8 & 73.1 & 59.4 & 35.7 & 59.6 & 72.9 \\
				& \textbf{RT DETR-R101~\cite{lv2023detrs} (I)} & 76 & 259 & 54.3 & 72.7 & 58.6 & 36.0 & 58.8 & 72.1 \\
				& \textbf{RTMDet-X~\cite{lyu2022rtmdet} (I)} & 94.9 & 283.4 & 52.8 & 70.4 & -- & -- & -- & -- \\
				& \textbf{Dy-YOLOv7-X~\cite{lin2023dynamicdet} (S)} & -- & 307.9 & 55.0 & 73.2 & 60.0 & 36.6 & 58.7 & 68.5 \\
				\bottomrule
			\end{tabular}		
			\begin{tablenotes}[flushleft]
				\footnotesize
				\item[1] (S), (I), (D), (C) indicate train-from-scratch, ImageNet pretrained, knowledge distillation, and complex setting, respectively.
			\end{tablenotes}
		\end{threeparttable}
	\end{table*}
	
	Shown in Table~\ref{table:flops} is the performance of all participating models sorted by the amount of computation. Our proposed YOLOv9 is Pareto optimal in all models with different scales. Among models with more than 60 GFLOPs, only ELAN-based DAMO-YOLO and DETR-based RT DETR can rival the proposed YOLOv9. The above comparison results show that YOLOv9 has the most outstanding performance in the trade-off between computation complexity and accuracy.
	

{\small
	\begin{thebibliography}{10}\itemsep=-1pt
		
		\bibitem{bao2022beit}
		Hangbo Bao, Li Dong, Songhao Piao, and Furu Wei.
		\newblock {BEiT}: {BERT} pre-training of image transformers.
		\newblock In {\em International Conference on Learning Representations (ICLR)},
		2022.
		
		\bibitem{bochkovskiy2020yolov4}
		Alexey Bochkovskiy, Chien-Yao Wang, and Hong-Yuan~Mark Liao.
		\newblock {YOLOv4}: Optimal speed and accuracy of object detection.
		\newblock {\em arXiv preprint arXiv:2004.10934}, 2020.
		
		\bibitem{cai2022reversible}
		Yuxuan Cai, Yizhuang Zhou, Qi Han, Jianjian Sun, Xiangwen Kong, Jun Li, and
		Xiangyu Zhang.
		\newblock Reversible column networks.
		\newblock In {\em International Conference on Learning Representations (ICLR)},
		2023.
		
		\bibitem{carion2020end}
		Nicolas Carion, Francisco Massa, Gabriel Synnaeve, Nicolas Usunier, Alexander
		Kirillov, and Sergey Zagoruyko.
		\newblock End-to-end object detection with transformers.
		\newblock In {\em Proceedings of the European Conference on Computer Vision
			(ECCV)}, pages 213--229, 2020.
		
		\bibitem{chen2020ap}
		Kean Chen, Weiyao Lin, Jianguo Li, John See, Ji Wang, and Junni Zou.
		\newblock {AP}-loss for accurate one-stage object detection.
		\newblock {\em IEEE Transactions on Pattern Analysis and Machine Intelligence
			(TPAMI)}, 43(11):3782--3798, 2020.
		
		\bibitem{chen2022sdae}
		Yabo Chen, Yuchen Liu, Dongsheng Jiang, Xiaopeng Zhang, Wenrui Dai, Hongkai
		Xiong, and Qi Tian.
		\newblock {SdAE}: Self-distillated masked autoencoder.
		\newblock In {\em Proceedings of the European Conference on Computer Vision
			(ECCV)}, pages 108--124, 2022.
		
		\bibitem{chen2023yolo}
		Yuming Chen, Xinbin Yuan, Ruiqi Wu, Jiabao Wang, Qibin Hou, and Ming-Ming
		Cheng.
		\newblock {YOLO-MS}: rethinking multi-scale representation learning for
		real-time object detection.
		\newblock {\em arXiv preprint arXiv:2308.05480}, 2023.
		
		\bibitem{ding2022davit}
		Mingyu Ding, Bin Xiao, Noel Codella, Ping Luo, Jingdong Wang, and Lu Yuan.
		\newblock {DaVIT}: Dual attention vision transformers.
		\newblock In {\em Proceedings of the European Conference on Computer Vision
			(ECCV)}, pages 74--92, 2022.
		
		\bibitem{dosovitskiy2021image}
		Alexey Dosovitskiy, Lucas Beyer, Alexander Kolesnikov, Dirk Weissenborn,
		Xiaohua Zhai, Thomas Unterthiner, Mostafa Dehghani, Matthias Minderer, Georg
		Heigold, Sylvain Gelly, et~al.
		\newblock An image is worth 16x16 words: Transformers for image recognition at
		scale.
		\newblock In {\em International Conference on Learning Representations (ICLR)},
		2021.
		
		\bibitem{feng2021tood}
		Chengjian Feng, Yujie Zhong, Yu Gao, Matthew~R Scott, and Weilin Huang.
		\newblock {TOOD}: Task-aligned one-stage object detection.
		\newblock In {\em Proceedings of the IEEE/CVF International Conference on
			Computer Vision (ICCV)}, pages 3490--3499, 2021.
		
		\bibitem{gao2019res2net}
		Shang-Hua Gao, Ming-Ming Cheng, Kai Zhao, Xin-Yu Zhang, Ming-Hsuan Yang, and
		Philip Torr.
		\newblock {Res2Net}: A new multi-scale backbone architecture.
		\newblock {\em IEEE Transactions on Pattern Analysis and Machine Intelligence
			(TPAMI)}, 43(2):652--662, 2019.
		
		\bibitem{ge2021ota}
		Zheng Ge, Songtao Liu, Zeming Li, Osamu Yoshie, and Jian Sun.
		\newblock {OTA}: Optimal transport assignment for object detection.
		\newblock In {\em Proceedings of the IEEE/CVF Conference on Computer Vision and
			Pattern Recognition (CVPR)}, pages 303--312, 2021.
		
		\bibitem{ge2021yolox}
		Zheng Ge, Songtao Liu, Feng Wang, Zeming Li, and Jian Sun.
		\newblock {YOLOX}: Exceeding {YOLO} series in 2021.
		\newblock {\em arXiv preprint arXiv:2107.08430}, 2021.
		
		\bibitem{glenn2022yolov5}
		Jocher Glenn.
		\newblock {YOLOv5} release v7.0.
		\newblock \url{https://github.com/ultralytics/yolov5/releases/tag/v7.0}, 2022.
		
		\bibitem{glenn2024yolov8}
		Jocher Glenn.
		\newblock {YOLOv8} release v8.1.0.
		\newblock \url{https://github.com/ultralytics/ultralytics/releases/tag/v8.1.0},
		2024.
		
		\bibitem{gomez2017reversible}
		Aidan~N Gomez, Mengye Ren, Raquel Urtasun, and Roger~B Grosse.
		\newblock The reversible residual network: Backpropagation without storing
		activations.
		\newblock {\em Advances in Neural Information Processing Systems (NeurIPS)},
		2017.
		
		\bibitem{gu2023mamba}
		Albert Gu and Tri Dao.
		\newblock Mamba: Linear-time sequence modeling with selective state spaces.
		\newblock {\em arXiv preprint arXiv:2312.00752}, 2023.
		
		\bibitem{guo2020augfpn}
		Chaoxu Guo, Bin Fan, Qian Zhang, Shiming Xiang, and Chunhong Pan.
		\newblock {AugFPN}: Improving multi-scale feature learning for object
		detection.
		\newblock In {\em Proceedings of the IEEE/CVF Conference on Computer Vision and
			Pattern Recognition (CVPR)}, pages 12595--12604, 2020.
		
		\bibitem{han2023revcolv2}
		Qi Han, Yuxuan Cai, and Xiangyu Zhang.
		\newblock {RevColV2}: Exploring disentangled representations in masked image
		modeling.
		\newblock {\em Advances in Neural Information Processing Systems (NeurIPS)},
		2023.
		
		\bibitem{hayder2017boundary}
		Zeeshan Hayder, Xuming He, and Mathieu Salzmann.
		\newblock Boundary-aware instance segmentation.
		\newblock In {\em Proceedings of the IEEE/CVF Conference on Computer Vision and
			Pattern Recognition (CVPR)}, pages 5696--5704, 2017.
		
		\bibitem{he2016deep}
		Kaiming He, Xiangyu Zhang, Shaoqing Ren, and Jian Sun.
		\newblock Deep residual learning for image recognition.
		\newblock In {\em Proceedings of the IEEE/CVF Conference on Computer Vision and
			Pattern Recognition (CVPR)}, pages 770--778, 2016.
		
		\bibitem{he2016identity}
		Kaiming He, Xiangyu Zhang, Shaoqing Ren, and Jian Sun.
		\newblock Identity mappings in deep residual networks.
		\newblock In {\em Proceedings of the European Conference on Computer Vision
			(ECCV)}, pages 630--645. Springer, 2016.
		
		\bibitem{huang2017densely}
		Gao Huang, Zhuang Liu, Laurens Van Der~Maaten, and Kilian~Q Weinberger.
		\newblock Densely connected convolutional networks.
		\newblock In {\em Proceedings of the IEEE/CVF Conference on Computer Vision and
			Pattern Recognition (CVPR)}, pages 4700--4708, 2017.
		
		\bibitem{huang2022monodtr}
		Kuan-Chih Huang, Tsung-Han Wu, Hung-Ting Su, and Winston~H Hsu.
		\newblock {MonoDTR}: Monocular {3D} object detection with depth-aware
		transformer.
		\newblock In {\em Proceedings of the IEEE/CVF Conference on Computer Vision and
			Pattern Recognition (CVPR)}, pages 4012--4021, 2022.
		
		\bibitem{huang2023yolocs}
		Lin Huang, Weisheng Li, Linlin Shen, Haojie Fu, Xue Xiao, and Suihan Xiao.
		\newblock {YOLOCS}: Object detection based on dense channel compression for
		feature spatial solidification.
		\newblock {\em arXiv preprint arXiv:2305.04170}, 2023.
		
		\bibitem{jaegle2021perceiver}
		Andrew Jaegle, Felix Gimeno, Andy Brock, Oriol Vinyals, Andrew Zisserman, and
		Joao Carreira.
		\newblock Perceiver: General perception with iterative attention.
		\newblock In {\em International Conference on Machine Learning (ICML)}, pages
		4651--4664, 2021.
		
		\bibitem{kenton2019bert}
		Jacob Devlin Ming-Wei~Chang Kenton and Lee~Kristina Toutanova.
		\newblock {BERT}: Pre-training of deep bidirectional transformers for language
		understanding.
		\newblock In {\em Proceedings of NAACL-HLT}, volume~1, page~2, 2019.
		
		\bibitem{lee2015deeply}
		Chen-Yu Lee, Saining Xie, Patrick Gallagher, Zhengyou Zhang, and Zhuowen Tu.
		\newblock Deeply-supervised nets.
		\newblock In {\em Artificial Intelligence and Statistics}, pages 562--570,
		2015.
		
		\bibitem{levinshtein2020datnet}
		Alex Levinshtein, Alborz~Rezazadeh Sereshkeh, and Konstantinos Derpanis.
		\newblock {DATNet}: Dense auxiliary tasks for object detection.
		\newblock In {\em Proceedings of the IEEE/CVF Winter Conference on Applications
			of Computer Vision (WACV)}, pages 1419--1427, 2020.
		
		\bibitem{li2023yolov6}
		Chuyi Li, Lulu Li, Yifei Geng, Hongliang Jiang, Meng Cheng, Bo Zhang, Zaidan
		Ke, Xiaoming Xu, and Xiangxiang Chu.
		\newblock {YOLOv6 v3.0}: A full-scale reloading.
		\newblock {\em arXiv preprint arXiv:2301.05586}, 2023.
		
		\bibitem{li2022yolov6}
		Chuyi Li, Lulu Li, Hongliang Jiang, Kaiheng Weng, Yifei Geng, Liang Li, Zaidan
		Ke, Qingyuan Li, Meng Cheng, Weiqiang Nie, et~al.
		\newblock {YOLOv6}: A single-stage object detection framework for industrial
		applications.
		\newblock {\em arXiv preprint arXiv:2209.02976}, 2022.
		
		\bibitem{li2023uni}
		Hao Li, Jinguo Zhu, Xiaohu Jiang, Xizhou Zhu, Hongsheng Li, Chun Yuan, Xiaohua
		Wang, Yu Qiao, Xiaogang Wang, Wenhai Wang, et~al.
		\newblock Uni-perceiver v2: A generalist model for large-scale vision and
		vision-language tasks.
		\newblock In {\em Proceedings of the IEEE/CVF Conference on Computer Vision and
			Pattern Recognition (CVPR)}, pages 2691--2700, 2023.
		
		\bibitem{li2022dual}
		Shuai Li, Chenhang He, Ruihuang Li, and Lei Zhang.
		\newblock A dual weighting label assignment scheme for object detection.
		\newblock In {\em Proceedings of the IEEE/CVF Conference on Computer Vision and
			Pattern Recognition (CVPR)}, pages 9387--9396, 2022.
		
		\bibitem{liang2021cbnetv2}
		Tingting Liang, Xiaojie Chu, Yudong Liu, Yongtao Wang, Zhi Tang, Wei Chu,
		Jingdong Chen, and Haibin Ling.
		\newblock {CBNet}: A composite backbone network architecture for object
		detection.
		\newblock {\em IEEE Transactions on Image Processing (TIP)}, 2022.
		
		\bibitem{lin2017feature}
		Tsung-Yi Lin, Piotr Doll{\'a}r, Ross Girshick, Kaiming He, Bharath Hariharan,
		and Serge Belongie.
		\newblock Feature pyramid networks for object detection.
		\newblock In {\em Proceedings of the IEEE/CVF Conference on Computer Vision and
			Pattern Recognition (CVPR)}, pages 2117--2125, 2017.
		
		\bibitem{lin2023dynamicdet}
		Zhihao Lin, Yongtao Wang, Jinhe Zhang, and Xiaojie Chu.
		\newblock {DynamicDet}: A unified dynamic architecture for object detection.
		\newblock In {\em Proceedings of the IEEE/CVF Conference on Computer Vision and
			Pattern Recognition (CVPR)}, pages 6282--6291, 2023.
		
		\bibitem{liu2018path}
		Shu Liu, Lu Qi, Haifang Qin, Jianping Shi, and Jiaya Jia.
		\newblock Path aggregation network for instance segmentation.
		\newblock In {\em Proceedings of the IEEE/CVF Conference on Computer Vision and
			Pattern Recognition (CVPR)}, pages 8759--8768, 2018.
		
		\bibitem{liu2024vmamba}
		Yue Liu, Yunjie Tian, Yuzhong Zhao, Hongtian Yu, Lingxi Xie, Yaowei Wang,
		Qixiang Ye, and Yunfan Liu.
		\newblock Vmamba: Visual state space model.
		\newblock {\em arXiv preprint arXiv:2401.10166}, 2024.
		
		\bibitem{liu2020cbnet}
		Yudong Liu, Yongtao Wang, Siwei Wang, TingTing Liang, Qijie Zhao, Zhi Tang, and
		Haibin Ling.
		\newblock {CBNet}: A novel composite backbone network architecture for object
		detection.
		\newblock In {\em Proceedings of the AAAI Conference on Artificial Intelligence
			(AAAI)}, pages 11653--11660, 2020.
		
		\bibitem{liu2022swin}
		Ze Liu, Han Hu, Yutong Lin, Zhuliang Yao, Zhenda Xie, Yixuan Wei, Jia Ning, Yue
		Cao, Zheng Zhang, Li Dong, et~al.
		\newblock Swin transformer v2: Scaling up capacity and resolution.
		\newblock In {\em Proceedings of the IEEE/CVF Conference on Computer Vision and
			Pattern Recognition (CVPR)}, 2022.
		
		\bibitem{liu2021swin}
		Ze Liu, Yutong Lin, Yue Cao, Han Hu, Yixuan Wei, Zheng Zhang, Stephen Lin, and
		Baining Guo.
		\newblock Swin transformer: Hierarchical vision transformer using shifted
		windows.
		\newblock In {\em Proceedings of the IEEE/CVF International Conference on
			Computer Vision (ICCV)}, pages 10012--10022, 2021.
		
		\bibitem{liu2022convnext}
		Zhuang Liu, Hanzi Mao, Chao-Yuan Wu, Christoph Feichtenhofer, Trevor Darrell,
		and Saining Xie.
		\newblock A {ConvNet} for the 2020s.
		\newblock In {\em Proceedings of the IEEE/CVF Conference on Computer Vision and
			Pattern Recognition (CVPR)}, pages 11976--11986, 2022.
		
		\bibitem{lv2023detrs}
		Wenyu Lv, Shangliang Xu, Yian Zhao, Guanzhong Wang, Jinman Wei, Cheng Cui,
		Yuning Du, Qingqing Dang, and Yi Liu.
		\newblock {DETRs} beat {YOLOs} on real-time object detection.
		\newblock {\em arXiv preprint arXiv:2304.08069}, 2023.
		
		\bibitem{lyu2022rtmdet}
		Chengqi Lyu, Wenwei Zhang, Haian Huang, Yue Zhou, Yudong Wang, Yanyi Liu,
		Shilong Zhang, and Kai Chen.
		\newblock {RTMDet}: An empirical study of designing real-time object detectors.
		\newblock {\em arXiv preprint arXiv:2212.07784}, 2022.
		
		\bibitem{oksuz2020ranking}
		Kemal Oksuz, Baris~Can Cam, Emre Akbas, and Sinan Kalkan.
		\newblock A ranking-based, balanced loss function unifying classification and
		localisation in object detection.
		\newblock {\em Advances in Neural Information Processing Systems (NeurIPS)},
		33:15534--15545, 2020.
		
		\bibitem{oksuz2021rank}
		Kemal Oksuz, Baris~Can Cam, Emre Akbas, and Sinan Kalkan.
		\newblock Rank \& sort loss for object detection and instance segmentation.
		\newblock In {\em Proceedings of the IEEE/CVF International Conference on
			Computer Vision (ICCV)}, pages 3009--3018, 2021.
		
		\bibitem{redmon2016you}
		Joseph Redmon, Santosh Divvala, Ross Girshick, and Ali Farhadi.
		\newblock You only look once: Unified, real-time object detection.
		\newblock In {\em Proceedings of the IEEE/CVF Conference on Computer Vision and
			Pattern Recognition (CVPR)}, pages 779--788, 2016.
		
		\bibitem{redmon2017yolo9000}
		Joseph Redmon and Ali Farhadi.
		\newblock {YOLO9000}: better, faster, stronger.
		\newblock In {\em Proceedings of the IEEE/CVF Conference on Computer Vision and
			Pattern Recognition (CVPR)}, pages 7263--7271, 2017.
		
		\bibitem{redmon2018yolov3}
		Joseph Redmon and Ali Farhadi.
		\newblock {YOLOv3}: An incremental improvement.
		\newblock {\em arXiv preprint arXiv:1804.02767}, 2018.
		
		\bibitem{rezatofighi2019generalized}
		Hamid Rezatofighi, Nathan Tsoi, JunYoung Gwak, Amir Sadeghian, Ian Reid, and
		Silvio Savarese.
		\newblock Generalized intersection over union: A metric and a loss for bounding
		box regression.
		\newblock In {\em Proceedings of the IEEE/CVF Conference on Computer Vision and
			Pattern Recognition (CVPR)}, pages 658--666, 2019.
		
		\bibitem{shen2019object}
		Zhiqiang Shen, Zhuang Liu, Jianguo Li, Yu-Gang Jiang, Yurong Chen, and
		Xiangyang Xue.
		\newblock Object detection from scratch with deep supervision.
		\newblock {\em IEEE Transactions on Pattern Analysis and Machine Intelligence
			(TPAMI)}, 42(2):398--412, 2019.
		
		\bibitem{shridhar2023perceiver}
		Mohit Shridhar, Lucas Manuelli, and Dieter Fox.
		\newblock Perceiver-actor: A multi-task transformer for robotic manipulation.
		\newblock In {\em Conference on Robot Learning (CoRL)}, pages 785--799, 2023.
		
		\bibitem{sun2021makes}
		Peize Sun, Yi Jiang, Enze Xie, Wenqi Shao, Zehuan Yuan, Changhu Wang, and Ping
		Luo.
		\newblock What makes for end-to-end object detection?
		\newblock In {\em International Conference on Machine Learning (ICML)}, pages
		9934--9944, 2021.
		
		\bibitem{szegedy2015going}
		Christian Szegedy, Wei Liu, Yangqing Jia, Pierre Sermanet, Scott Reed, Dragomir
		Anguelov, Dumitru Erhan, Vincent Vanhoucke, and Andrew Rabinovich.
		\newblock Going deeper with convolutions.
		\newblock In {\em Proceedings of the IEEE/CVF Conference on Computer Vision and
			Pattern Recognition (CVPR)}, pages 1--9, 2015.
		
		\bibitem{szegedy2016rethinking}
		Christian Szegedy, Vincent Vanhoucke, Sergey Ioffe, Jon Shlens, and Zbigniew
		Wojna.
		\newblock Rethinking the inception architecture for computer vision.
		\newblock In {\em Proceedings of the IEEE/CVF Conference on Computer Vision and
			Pattern Recognition (CVPR)}, pages 2818--2826, 2016.
		
		\bibitem{tang2023perceiver}
		Zineng Tang, Jaemin Cho, Jie Lei, and Mohit Bansal.
		\newblock {Perceiver-VL}: Efficient vision-and-language modeling with iterative
		latent attention.
		\newblock In {\em Proceedings of the IEEE/CVF Winter Conference on Applications
			of Computer Vision (WACV)}, pages 4410--4420, 2023.
		
		\bibitem{tian2019fcos}
		Zhi Tian, Chunhua Shen, Hao Chen, and Tong He.
		\newblock {FCOS}: Fully convolutional one-stage object detection.
		\newblock In {\em Proceedings of the IEEE/CVF International Conference on
			Computer Vision (ICCV)}, pages 9627--9636, 2019.
		
		\bibitem{tian2022fcos}
		Zhi Tian, Chunhua Shen, Hao Chen, and Tong He.
		\newblock {FCOS}: A simple and strong anchor-free object detector.
		\newblock {\em IEEE Transactions on Pattern Analysis and Machine Intelligence
			(TPAMI)}, 44(4):1922--1933, 2022.
		
		\bibitem{tishby2015deep}
		Naftali Tishby and Noga Zaslavsky.
		\newblock Deep learning and the information bottleneck principle.
		\newblock In {\em IEEE Information Theory Workshop (ITW)}, pages 1--5, 2015.
		
		\bibitem{tu2022maxvit}
		Zhengzhong Tu, Hossein Talebi, Han Zhang, Feng Yang, Peyman Milanfar, Alan
		Bovik, and Yinxiao Li.
		\newblock {MaxVIT}: Multi-axis vision transformer.
		\newblock In {\em Proceedings of the European Conference on Computer Vision
			(ECCV)}, pages 459--479, 2022.
		
		\bibitem{wang2023gold}
		Chengcheng Wang, Wei He, Ying Nie, Jianyuan Guo, Chuanjian Liu, Kai Han, and
		Yunhe Wang.
		\newblock {Gold-YOLO}: Efficient object detector via gather-and-distribute
		mechanism.
		\newblock {\em Advances in Neural Information Processing Systems (NeurIPS)},
		2023.
		
		\bibitem{wang2021scaled}
		Chien-Yao Wang, Alexey Bochkovskiy, and Hong-Yuan~Mark Liao.
		\newblock {Scaled-YOLOv4}: Scaling cross stage partial network.
		\newblock In {\em Proceedings of the IEEE/CVF Conference on Computer Vision and
			Pattern Recognition (CVPR)}, pages 13029--13038, 2021.
		
		\bibitem{wang2023yolov7}
		Chien-Yao Wang, Alexey Bochkovskiy, and Hong-Yuan~Mark Liao.
		\newblock {YOLOv7}: Trainable bag-of-freebies sets new state-of-the-art for
		real-time object detectors.
		\newblock In {\em Proceedings of the IEEE/CVF Conference on Computer Vision and
			Pattern Recognition (CVPR)}, pages 7464--7475, 2023.
		
		\bibitem{wang2020cspnet}
		Chien-Yao Wang, Hong-Yuan~Mark Liao, Yueh-Hua Wu, Ping-Yang Chen, Jun-Wei
		Hsieh, and I-Hau Yeh.
		\newblock {CSPNet}: A new backbone that can enhance learning capability of
		{CNN}.
		\newblock In {\em Proceedings of the IEEE/CVF Conference on Computer Vision and
			Pattern Recognition Workshops (CVPRW)}, pages 390--391, 2020.
		
		\bibitem{wang2023designing}
		Chien-Yao Wang, Hong-Yuan~Mark Liao, and I-Hau Yeh.
		\newblock Designing network design strategies through gradient path analysis.
		\newblock {\em Journal of Information Science and Engineering (JISE)},
		39(4):975--995, 2023.
		
		\bibitem{wang2021you}
		Chien-Yao Wang, I-Hau Yeh, and Hong-Yuan~Mark Liao.
		\newblock You only learn one representation: Unified network for multiple
		tasks.
		\newblock {\em Journal of Information Science \& Engineering (JISE)},
		39(3):691--709, 2023.
		
		\bibitem{wang2021end}
		Jianfeng Wang, Lin Song, Zeming Li, Hongbin Sun, Jian Sun, and Nanning Zheng.
		\newblock End-to-end object detection with fully convolutional network.
		\newblock In {\em Proceedings of the IEEE/CVF Conference on Computer Vision and
			Pattern Recognition (CVPR)}, pages 15849--15858, 2021.
		
		\bibitem{wang2015training}
		Liwei Wang, Chen-Yu Lee, Zhuowen Tu, and Svetlana Lazebnik.
		\newblock Training deeper convolutional networks with deep supervision.
		\newblock {\em arXiv preprint arXiv:1505.02496}, 2015.
		
		\bibitem{wang2021pyramid}
		Wenhai Wang, Enze Xie, Xiang Li, Deng-Ping Fan, Kaitao Song, Ding Liang, Tong
		Lu, Ping Luo, and Ling Shao.
		\newblock Pyramid vision transformer: A versatile backbone for dense prediction
		without convolutions.
		\newblock In {\em Proceedings of the IEEE/CVF International Conference on
			Computer Vision (ICCV)}, pages 568--578, 2021.
		
		\bibitem{wang2022pvt}
		Wenhai Wang, Enze Xie, Xiang Li, Deng-Ping Fan, Kaitao Song, Ding Liang, Tong
		Lu, Ping Luo, and Ling Shao.
		\newblock {PVT v2}: Improved baselines with pyramid vision transformer.
		\newblock {\em Computational Visual Media}, 8(3):415--424, 2022.
		
		\bibitem{woo2023convnext}
		Sanghyun Woo, Shoubhik Debnath, Ronghang Hu, Xinlei Chen, Zhuang Liu, In~So
		Kweon, and Saining Xie.
		\newblock {ConvNeXt v2}: Co-designing and scaling convnets with masked
		autoencoders.
		\newblock In {\em Proceedings of the IEEE/CVF Conference on Computer Vision and
			Pattern Recognition (CVPR)}, pages 16133--16142, 2023.
		
		\bibitem{xie2017aggregated}
		Saining Xie, Ross Girshick, Piotr Doll{\'a}r, Zhuowen Tu, and Kaiming He.
		\newblock Aggregated residual transformations for deep neural networks.
		\newblock In {\em Proceedings of the IEEE/CVF Conference on Computer Vision and
			Pattern Recognition (CVPR)}, pages 1492--1500, 2017.
		
		\bibitem{xie2022simmim}
		Zhenda Xie, Zheng Zhang, Yue Cao, Yutong Lin, Jianmin Bao, Zhuliang Yao, Qi
		Dai, and Han Hu.
		\newblock {SimMIM}: A simple framework for masked image modeling.
		\newblock In {\em Proceedings of the IEEE/CVF Conference on Computer Vision and
			Pattern Recognition (CVPR)}, pages 9653--9663, 2022.
		
		\bibitem{xu2022pp}
		Shangliang Xu, Xinxin Wang, Wenyu Lv, Qinyao Chang, Cheng Cui, Kaipeng Deng,
		Guanzhong Wang, Qingqing Dang, Shengyu Wei, Yuning Du, et~al.
		\newblock {PP-YOLOE}: An evolved version of {YOLO}.
		\newblock {\em arXiv preprint arXiv:2203.16250}, 2022.
		
		\bibitem{xu2022damo}
		Xianzhe Xu, Yiqi Jiang, Weihua Chen, Yilun Huang, Yuan Zhang, and Xiuyu Sun.
		\newblock {DAMO-YOLO}: A report on real-time object detection design.
		\newblock {\em arXiv preprint arXiv:2211.15444}, 2022.
		
		\bibitem{zhang2023monodetr}
		Renrui Zhang, Han Qiu, Tai Wang, Ziyu Guo, Ziteng Cui, Yu Qiao, Hongsheng Li,
		and Peng Gao.
		\newblock {MonoDETR}: Depth-guided transformer for monocular {3D} object
		detection.
		\newblock In {\em Proceedings of the IEEE/CVF International Conference on
			Computer Vision (ICCV)}, pages 9155--9166, 2023.
		
		\bibitem{zheng2020distance}
		Zhaohui Zheng, Ping Wang, Wei Liu, Jinze Li, Rongguang Ye, and Dongwei Ren.
		\newblock {Distance-IoU} loss: Faster and better learning for bounding box
		regression.
		\newblock In {\em Proceedings of the AAAI Conference on Artificial Intelligence
			(AAAI)}, volume~34, pages 12993--13000, 2020.
		
		\bibitem{zhou2019iou}
		Dingfu Zhou, Jin Fang, Xibin Song, Chenye Guan, Junbo Yin, Yuchao Dai, and
		Ruigang Yang.
		\newblock {IoU} loss for {2D}/{3D} object detection.
		\newblock In {\em International Conference on 3D Vision (3DV)}, pages 85--94,
		2019.
		
		\bibitem{zhu2020autoassign}
		Benjin Zhu, Jianfeng Wang, Zhengkai Jiang, Fuhang Zong, Songtao Liu, Zeming Li,
		and Jian Sun.
		\newblock {AutoAssign}: Differentiable label assignment for dense object
		detection.
		\newblock {\em arXiv preprint arXiv:2007.03496}, 2020.
		
		\bibitem{zhu2024vision}
		Lianghui Zhu, Bencheng Liao, Qian Zhang, Xinlong Wang, Wenyu Liu, and Xinggang
		Wang.
		\newblock Vision mamba: Efficient visual representation learning with
		bidirectional state space model.
		\newblock {\em arXiv preprint arXiv:2401.09417}, 2024.
		
		\bibitem{zhu2022uni}
		Xizhou Zhu, Jinguo Zhu, Hao Li, Xiaoshi Wu, Hongsheng Li, Xiaohua Wang, and
		Jifeng Dai.
		\newblock Uni-perceiver: Pre-training unified architecture for generic
		perception for zero-shot and few-shot tasks.
		\newblock In {\em Proceedings of the IEEE/CVF Conference on Computer Vision and
			Pattern Recognition (CVPR)}, pages 16804--16815, 2022.
		
		\bibitem{zong2023detrs}
		Zhuofan Zong, Guanglu Song, and Yu Liu.
		\newblock {DETRs} with collaborative hybrid assignments training.
		\newblock In {\em Proceedings of the IEEE/CVF Conference on Computer Vision and
			Pattern Recognition (CVPR)}, pages 6748--6758, 2023.
		
	\end{thebibliography}
	
	}
\end{document}